\documentclass[10pt,twocolumn,letterpaper]{article}

\usepackage[pagenumbers]{cvpr}      

\usepackage{booktabs}
\usepackage{graphicx}
\usepackage{amsmath,amssymb}
\usepackage{xcolor}

\definecolor{cvprblue}{rgb}{0.21,0.49,0.74}
\usepackage[pagebackref,breaklinks,colorlinks,allcolors=cvprblue]{hyperref}

\def\confName{CVPR}
\def\confYear{2027}

\title{Option Order and Labels in Jev-Style Visual Decisions:\\ What Training-Free Debiasing Transfers from Text to Vision}

\author{Youzhi Zhang\\
 Hong Kong Institute of Science \& Innovation, Chinese Academy of Sciences}

\begin{document}
\maketitle
\begin{abstract}
Jev-style decision models (typed system-1 interfaces) read a choice from one prefill
and require no generation, parsing or training. On text, two training-free
corrections are standard: average over cyclic rotations of the option list (L0)
and divide out a label prior. We ask whether they transfer to vision-language
models. We evaluate seven VLMs from four families; we measure
listing-order sensitivity, label bias, sorting, and pointwise scoring. Position bias is large:
reversing the list changes 3.5--36\% of answers on the non-MMMU datasets and
30--45\% on MMMU. 
Overall, L0 reduces the reversal flip in every full cell;
accuracy gains are family-dependent. Pointwise scoring usually loses accuracy, which is different from the results on text, while the reduced cells agree on flips.
Debiasing transfers when the bias is identifiable, not by default.  Our code and data are publicly available at
\url{https://github.com/Youzhi333/Option_Order_Label_Visual_Jev}.
\end{abstract}

\section{Introduction}
\label{sec:intro}

Jev-style decision models are typed system-1 interfaces for choice \citep{typesafe2026jev,almeida2026jev}. Instead of
generating a rationale and parsing an answer, they expose a typed decision over
a supplied option set and read it from one forward pass. AnyJev ~\citep{anyjev2026} turns an
instruction-tuned language model into such a model by restricting its
next-token distribution to the option labels; the result is fast, parse-free,
and carries a probability for every option. Visual Jev
applies the same interface to a shared visual context~\citep{yu2026visualjev},
and PixelJev maps an image, an instruction and a runtime candidate set to a
structured choice with candidate-conditioned probabilities~\citep{zhou2026pixeljev}.
The appeal is operational: one prefill, no generation, and no parser.

A visual Jev inherits the failure modes of its readout, not just its speed. On
text, the answer depends on option order~\citep{zheng2024large,
pezeshkpour-hruschka-2024-large}, and the model prefers some labels regardless
of the input~\citep{zhao2021calibrate,zhou2024batch}. Two training-free
corrections are standard: read every cyclic rotation and average in log space
(L0), and divide out a label prior estimated without labels. They are cheap, and
on text they lower the order-flip rate from roughly a third to a sixth while
raising accuracy~\citep{anyjev2026}. It is tempting to assume they transfer to
images unchanged. Recent work shows that order sensitivity is also large in
multimodal LLMs
\citep{tan2024order,wei2026order,zong2024fool,paruchuri2026same,zhu2026stable,rosenthal2026unexplored}
and that LVLMs carry answer priors
\citep{zhang2024debiasing,roh2026choices,modi2026readout}, but it does not
separate a position effect from a class prior on a fixed option list, or ask
when a batch prior is estimable.

Vision changes the structure of the decision. The option set can be
\emph{fixed} across items, as in a recognition benchmark where the same class
list is shown for every image, or \emph{per-item}, as in visual question
answering. A fixed list turns position into an alias for class: slot 0 is
always annual crop, slot 5 always pasture. A per-item list makes the batch
prior of text benchmarks impossible to estimate, because each question is
asked once. And because a VLM can be asked about one option at a time, there is
a third readout with no counterpart in the letter readout: score each option
alone and ask whether it is correct (pointwise)~\citep{huang2026runoff}.

\begin{figure*}[t]
  \centering
  \includegraphics[width=\textwidth]{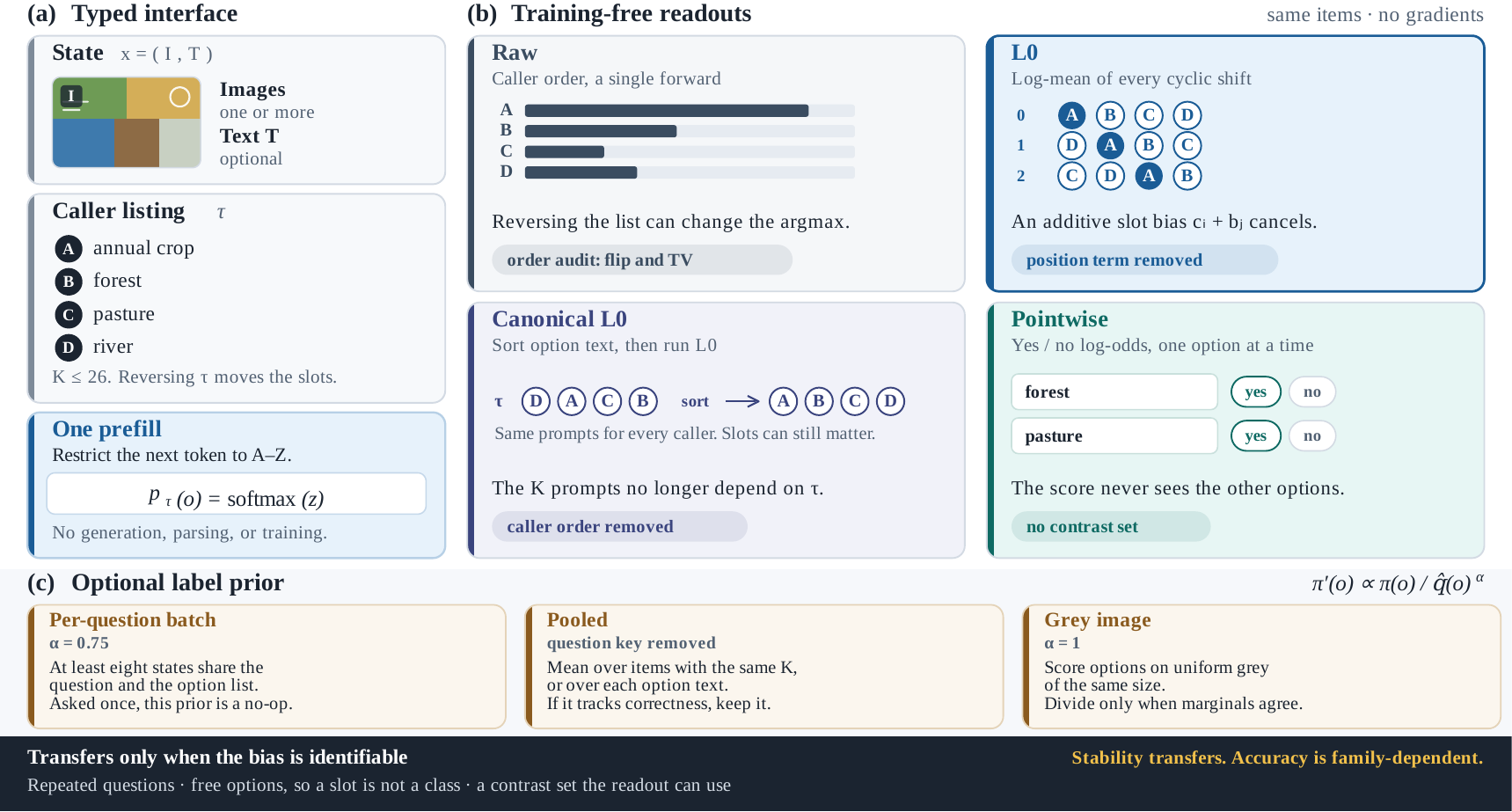}
  \caption{Training-free framework for a Jev-style visual decision.
  (a)~A state, a question, and a caller listing are scored in one prefill by restricting the next-token distribution to the option labels.
  (b)~Raw uses that listing once. L0 log-averages every cyclic shift and cancels an additive slot bias. Canonical L0 sorts option text first, so the prompts no longer depend on the caller. Pointwise scores each option alone and therefore has no contrast set.
  (c)~A label prior is divided out only when it estimates bias rather than the answer.}
  \label{fig:framework}
\end{figure*}

To study the properties of Jev-style visual decisions, as shown in Figure \ref{fig:framework},
we study the option-list axis: whether the text corrections transfer when the
state is an image. We do not study visual-versus-text token
order~\citep{gan2026order,wei2026order} or re-evaluate text-only bias; both are
documented, and our intervention reorders options across slots, not modality
tokens. Throughout, ``system-1'' denotes this typed, label-restricted readout;
we do not test generative multiple-choice answering, chain-of-thought, or
free-form answers. We do not train or fine-tune.

We study seven VLMs from four families: a four-model Qwen \citep{qwen3vl2025,wang2024qwen2vl} grid on eight
datasets, 12 full-cell cross-family runs of
LLaVA-OneVision \citep{li2024llavaonevision}, LLaVA-NeXT \citep{liu2024llavanext} and Idefics3 \citep{laurencon2024idefics3} spanning A-OKVQA~\citep{schwenk2022aokvqa}, ScienceQA~\citep{lu2022scienceqa}, EuroSAT~\citep{helber2019eurosat}
and Pets~\citep{parkhi2012pets}, with LLaVA-NeXT receiving the full method protocol. Option counts range from $K{=}2$--26. We measure reversal flip,   the aligned total-variation
distance (TV), L0's change on a fixed
listing, calibration, and paired bootstrap intervals. We compare caller-order
L0, canonical (text-sorted) L0, pointwise scoring, and per-question, pooled and
grey priors. 

Our findings are mostly negative, and that is the point.

\begin{itemize}
\item \textbf{Position bias transfers.} Reversal changes 3.5--36\% of answers
on the non-MMMU sets (30--45\% on MMMU); in the Qwen grid, L0 cuts the flip
by 1.1--6.6$\times$ and improves the point estimate in most cells.
\item \textbf{But L0 is not universally safe.} On EuroSAT 4B the point estimate
loses 2.8 points, but the class-cluster interval includes zero. The class list
is fixed, so position is a class preference, and raw beats L0 on high-probability
classes ($r{=}0.64$).
\item \textbf{Sorting removes listing ambiguity, not slot bias.} The sorted L0
has zero reversal flip except for exact ties (45 of 32,400 item-reads, 0.14\%;
at most 0.95\% in a cell),
but its single read and cycle average still differ by mean TV up to 0.22.
\item \textbf{The label prior needs repeated questions.} It is a no-op on
A-OKVQA, the synthetic $K{=}8/16$ sets and MMMU; on Pets and EuroSAT, it gains
0.3--4.5 points. On open-set VQA the pooled prior correlates $+0.83$ to $+0.86$
with correctness, so dividing it removes signal.
\item \textbf{Pointwise is not a drop-in replacement.} On the Qwen grid it beats
L0 only on Pets (+0.12; 0.901 with the batch prior); it loses 1.4--4.2 on non-Pets VQA sets and 5.8 on MMMU, more with $K$. The trend is different from the results on text \citep{huang2026runoff,zhang2026ijev}. 
\item \textbf{A content-free prior works only when aligned.} The grey prior
gains +10.5 points on EuroSAT 2B at $\alpha{=}1$, where it and the real
marginal both over-predict brushland, and hurts when they disagree.
\end{itemize}

The practical conclusion is a checklist rather than a method: measure the
reversal flip, check whether the prior is estimable, correlate the prior with
label correctness before dividing it, and treat a fixed option list as a
confound between position and class. Section~\ref{sec:analysis} turns these
observations into diagnostics.

\section{Related Work}
\label{sec:related}

\paragraph{Typed decisions from one prefill.}
AnyJev restricts the next-token distribution of an instruction-tuned language
model to the option labels and reads a decision from one prefill; it corrects
label and position bias without gradients~\citep{anyjev2026}. Visual Jev moves
the same idea to images, shares the visual prefix across independent questions
and shows that answer-supervised post-training, not a typed head, drives
accuracy~\citep{yu2026visualjev}. PixelJev argues for a native-image decision
interface over runtime candidate sets and evaluates frozen and adapted small
multimodal models~\citep{zhou2026pixeljev}. We use the readout these systems share and train nothing.

\paragraph{Order sensitivity.}
Option order changes multiple-choice answers in language models, and the effect
grows with the number of options~\citep{zheng2024large,
pezeshkpour-hruschka-2024-large}. Cyclic permutation marginalization (L0)
averages log-probabilities over rotations and removes an additive position
effect exactly~\citep{zheng2024large,anyjev2026}. Sorting the options before
rotating removes the remaining dependence on which options are adjacent, making
the prompt a function of the option set~\citep{anyjev2026}. Order-independence
has been approached by calibration~\citep{mcilroyyoung2024orderindependence}, by
scoring each option alone, and by combining pointwise scores with a
position-balanced runoff~\citep{huang2026runoff,zhang2026ijev}. The effect is now also
documented in vision-language models. \citet{wei2026order} evaluate choice and
modality ordering on MMMU, CVQA and MMBench across 14 LVLMs and find both
matter, with text-side mitigation not transferring cleanly.
\citet{tan2024order} find that MLLMs
prefer the beginning and end of a multimodal context, with accuracy highest when
the correct option is shown first or last. \citet{paruchuri2026same} audit 18 MLLMs across five ordering facets and report
24--50\% per-facet flips (13.4\% for the best model), with prompt-level
mitigation not transferring from text; \citet{zhu2026stable} report
40.8--66.4\% per-cell flips in video multiple-choice while the score moves by at
most 5.8 points. \citet{rosenthal2026unexplored} vary option IDs,
delimiters and separators across 48 formats and find rank reversals, tracing
part of the instability to tokenizer-level option-ID corruption.
\citet{atabuzzaman2025selection} characterize MCQA selection bias in LVLMs and
propose an inference-time logit correction. \citet{zong2024fool} show that
permuting answer options is an embarrassingly simple attack that drops MCQA
accuracy sharply for LLMs and VLLMs, and that calibration does not fix it.
\citet{gan2026order} and \citet{zhang2026visiontextbiasjev} study the order of visual and text tokens rather than
options: visual tokens placed last underperform those placed first, and a
dual-order contrastive decoder narrows the gap. These studies establish that the
effect is large. They do not separate a position effect from a class prior on a
fixed option list in Jev-style visual decisions, and they do not ask when a batch label prior is estimable.
Relative to these studies, we separate position from fixed-class priors,
validate prior estimability, and test sorting, pointwise and grey priors in Jev-style visual decisions.

\paragraph{Label priors in vision.}
Classic VQA documents language and answer priors: VQA-CP rebalances the answer
distribution and strong models lose accuracy on it~\citep{agrawal2018dont}, and
RUBi reduces unimodal biases~\citep{cadene2019rubi}. In LVLMs,
\citet{zhang2024debiasing} show that models answer from the underlying LLM prior
even when the image is absent or incongruent, and propose a post-hoc affine
calibration for classification and multiple-choice plus a contrastive
visual-debias decoding. \citet{roh2026choices} show that the answer choices
themselves induce a prior and subtract a text-only logit in a training-free
contrastive decoder. \citet{modi2026readout} show that the answer readout is
itself a major confounder in medical MC-VQA: switching from isolated
option-text likelihood to joint option presentation with symbol scoring raises
accuracy from 0.643 to 0.954 with unchanged weights (their comparison bundles
several changes), while a shortest-option baseline reaches 0.899. Beyond the readout, \citet{vo2026biased} show VLMs rely on memorized
knowledge, \citet{luo2025probing} expose visual-language priors with ViLP, and
\citet{lin2024revisiting} propose test-time $\alpha$-debiasing for generative
scoring. Our grey-image probe is the readout-time analogue of the
image-free calibration of \citet{zhang2024debiasing}; we test it at the decision
level and show that it helps only when the content-free and real marginals align.
Unlike contrastive-decoding and logit-correction methods
\citep{zhang2024debiasing,roh2026choices,atabuzzaman2025selection}, which
subtract a prior at inference, we ask whether the prior is estimable and whether
dividing it removes signal in Jev-style visual decisions; the answer is dataset-dependent.

\paragraph{Label priors and calibration in text.}
A content-free prior can be estimated from probe inputs and divided out
\citep{zhao2021calibrate}; a batch prior estimates it from the inputs being
decided~\citep{zhou2024batch}. Both assume that the model's label marginal is a
bias rather than signal, and the batch variant assumes the batch is not
dominated by one label~\citep{anyjev2026}. Calibration of neural
networks~\citep{guo2017calibration} and selective
prediction~\citep{geifman2017selective} are the downstream concerns; we report
accuracy with paired bootstrap confidence intervals~\citep{efron1979bootstrap}.
Our results delimit when the assumption holds in Jev-style visual decisions.


\section{Readouts, Priors and Metrics}
\label{sec:method}

\paragraph{Setup.}
A state $x=(I,T)$ is one or more images $I$ with optional text $T$. A question
$q$ comes with an option set $\mathcal{O}=\{o_1,\dots,o_K\}$; a caller supplies
an ordered listing $\tau=(\tau_1,\dots,\tau_K)$ of those options. We render the
images, the question and the listed options with labels
$\ell_1,\dots,\ell_K\in\{\mathrm{A},\dots,\mathrm{Z}\}$, and ask for the label
only. We use $K\le26$ and single-token labels. One prefill gives logits $z_1,\dots,z_K$ at the answer position. The
readout is
\begin{equation}
  p_\tau(o_{\tau_j}) = \frac{e^{z_j}}{\sum_{k=1}^{K} e^{z_k}},
  \label{eq:restrict}
\end{equation}
which is a distribution over the option texts; nothing is generated and nothing is
parsed. The subscript $\tau$ records the listing: the same option can occupy a
different slot, and receive a different $z_j$, under a different listing.
\emph{raw} is $p^{\mathrm{raw}}=p_{\tau_{\mathrm{caller}}}$. Unless stated
otherwise, operations compare distributions after aligning them by option text,
not by slot.

\paragraph{L0.}
For $t=0,\dots,K-1$, let $\pi_t(j)=(j+t)\bmod K$. Listing shift $t$ places
option $\tau_i$ at slot $\pi_t^{-1}(i)$; reading that prompt gives
$p_t(j)$. Reindexing to options, $q_t(i)=p_t(\pi_t^{-1}(i))$. L0 \cite{anyjev2026} averages in log space,
\begin{equation}
  \bar z_i = \frac{1}{K}\sum_{t=0}^{K-1}\log q_t(i),
  \qquad
  p^{\mathrm{L0}} = \operatorname{softmax}(\bar z).
  \label{eq:l0}
\end{equation}
If the logit of option $i$ at slot $j$ separates as $c_i+b_j$, then
$\bar z_i = c_i + \frac{1}{K}\sum_j b_j - \frac{1}{K}\sum_t \log Z_t$, and the
last two terms do not depend on $i$: $\operatorname{softmax}(\bar z) =
\operatorname{softmax}(c)$. Under this additive model the position effect cancels exactly; empirically the
cancellation is only approximate. Averaging
probabilities instead of log-probabilities does not have this property.

\paragraph{Canonical order.}
L0 removes the position effect but not the effect of which options are adjacent.
The canonical operator $S(\mathcal{O})$ sorts the options by their text,
$\operatorname{str}(o)$, with the caller index as a tie-break. Our loaders
deduplicate option strings, so the key is unique: $S(\mathcal{O})$ is the same
sequence for \emph{any} caller order, and its $K$ cyclic listings are the same
$K$ prompts. L0 on $S(\mathcal{O})$ is therefore invariant to the caller's
order. The raw level still uses the caller's listing, so raw versus reversal remains a
measurement. The residual reversals in the canonical grid
(Table~\ref{tab:app-canonical}) are exact probability ties: the argmax indexes
the caller's array, and the same index names a different option under reversal
(Section~\ref{sec:canonical}).

\paragraph{Pointwise.}
For each option $o$ alone, ask whether it is the correct answer and read the
yes/no log-odds \cite{huang2026runoff,zhang2026ijev},
\begin{equation}
\begin{aligned}
  f(x,q,o) &= \log P(\text{yes}\mid x,q,o) - \log P(\text{no}\mid x,q,o),\\
  \pi(o) &= \frac{e^{f(x,q,o)}}{\sum_{o'\in\mathcal{O}} e^{f(x,q,o')}}.
\end{aligned}
\label{eq:pointwise}
\end{equation}
The prompt shows the image, then ``Question: \ldots Option: $o$. Is this option
the correct answer? Answer yes or no.''; yes and no are read as single tokens,
including capitalised and leading-space forms. No option is seen next to
another, so $\pi$ is listing-order invariant by construction. The base readout
applies no prior.

\paragraph{Label priors.}
A prior $\hat q$ is divided out at the decision layer,
\begin{equation}
  \pi'(o) \;\propto\; \pi(o)\,/\,\hat q(o)^{\alpha},
  \label{eq:prior}
\end{equation}
with $\alpha=0$ no prior, $\alpha=0.75$ the batch-prior default of AnyJev
0.3.0~\citep{anyjev2026}, and $\alpha=1$ the content-free default. For L0 the division is applied to each shift's position
distribution before the log-average in Equation~(\ref{eq:l0}); for pointwise it is
applied to $\pi$ in Equation~(\ref{eq:pointwise}). Three estimators:
\begin{itemize}
\item \emph{per-question batch.} $\hat q(o)=\frac1M\sum_{m=1}^{M}p_m(o)$ over
the $M\ge8$ states with the same question and option list seen so far; it is
undefined when a question occurs once. The current item is included, so this
estimate is transductive; deployment would need a held-out prefix.
\item \emph{pooled.} $\hat q$ is the mean over every item with the same $K$ (for
L0), or the mean $\pi$ of each option text over the items on which it appears
(for pointwise). This is the batch prior with the per-question key removed.
\item \emph{content-free grey.} Replace every image by uniform grey of the same
size, score each option alone, and let $\hat q=\pi_{\mathrm{grey}}$. In
log-odds form the correction is
$\pi_\alpha(o)=\operatorname{softmax}_o[f_{\mathrm{real}}(o)-\alpha
f_{\mathrm{grey}}(o)]$, the pointwise analogue of a content-free
prior~\citep{zhao2021calibrate}.
\end{itemize}

\section{Experimental Setup}
\label{sec:setup}

\paragraph{Datasets.} We use A-OKVQA~\citep{schwenk2022aokvqa}, ScienceQA~\citep{lu2022scienceqa},
MMMU~\citep{yue2024mmmu}, CV-Bench~\citep{tong2024cambrian},
EuroSAT~\citep{helber2019eurosat} and Oxford-IIIT Pets~\citep{parkhi2012pets},
with synthetic 8- and 16-option expansions of A-OKVQA. 
Table~\ref{tab:data} in the appendix lists the evaluation sets. ScienceQA and
CV-Bench are per-item VQA with short option lists; A-OKVQA $K{=}8/16$ are
synthetic expansions of the same 300 A-OKVQA items with distractors drawn from
the dataset's option pool, so their $K$ trend is a stress test, not a natural
large-$K$ benchmark; EuroSAT and Pets use a fixed option list for every
item; MMMU has 2--9 options and often several images. 

\paragraph{Models.} The models are
Qwen3-VL~\citep{qwen3vl2025}, Qwen2.5-VL~\citep{wang2024qwen2vl}, LLaVA-OneVision \citep{li2024llavaonevision}, LLaVA-NeXT \citep{liu2024llavanext}, and Idefics3 \citep{laurencon2024idefics3}. 
Qwen3-VL-2B/4B/8B-Instruct and Qwen2.5-VL-7B-Instruct, in bfloat16, with the
image token budget capped at $448^2$ pixels. Batch size is 4 for the 2B model
and 1 for the larger models. Runs are single-GPU (RTX 4090 D); four faulty
logical CPUs (8--11) are offlined. Result JSONs record the library versions,
GPU, batch size and dtype; the later full-cell runs also record the driver. Appendix \ref{sec:app_runtime} shows the runtime of these models.

\paragraph{Protocol.}
Each item is decided once per readout. The  position grid uses
\emph{no} label prior, so its L0 column isolates rotation averaging; priors are
measured in a separate runner on the same items. Reversal runs reverse the
caller's option list; canonical runs sort it by text inside L0. The pointwise
and grey runners reuse the same item selection. For the fixed-option datasets
we compare raw, caller L0, canonical L0, pointwise and the priors per cell.
Appendix~\ref{sec:appendix} gives prompts, hyperparameters and detailed results.

\paragraph{Metrics.}
Let $N$ be the number of items and $o^\star$ the gold option.
\emph{Accuracy} is $\frac{1}{N}\sum_n \mathbf{1}[\arg\max_o p_n(o)=o^\star_n]$.
\emph{ECE} is
$\sum_{b=1}^{10}\frac{n_b}{N}|\mathrm{acc}_b-\mathrm{conf}_b|$ over top-1
confidence bins, reported for the grey experiment. \emph{Coverage at 5\% risk} is the accepted fraction at the
most permissive threshold whose empirical selective risk is at most 5\%.
The \emph{reversal flip} is
$\frac{1}{N}\sum_n \mathbf{1}[\arg\max_o p_n(o)\neq
\arg\max_o p_n^{\mathrm{rev}}(o)]$, with $p^{\mathrm{rev}}$ aligned by option
text; it ignores how much probability moved. The \emph{reversal TV} is the mean
$\frac12\sum_o|p_n(o)-p_n^{\mathrm{rev}}(o)|$ after alignment, in $[0,1]$. The
\emph{correction TV} is the same distance between the single canonical read and
the cycle-averaged read on the same listing; it measures the slot bias that
sorting does not remove. Confidence intervals are percentile bootstraps over
items (2,000--4,000 resamples); for Pets and EuroSAT we also resample true
classes, since every item shares one option list.

\section{Results}
\label{sec:results}

\begin{table*}[t]
\centering
\footnotesize
\resizebox{\textwidth}{!}{%
\begin{tabular}{@{}lccccccccccccccc@{}}
\toprule
Dataset & Model & $n$ & $K$ & raw & L0 & L0+b & point & pw+b & grey & L0$_{\mathrm{can}}$ & $\Delta_{\mathrm{can}}$ & flip raw & flip L0 & flip L0$_{\mathrm{can}}$ & cTV \\
\midrule
ScienceQA & 2B & 2017 & 2--5 & 0.867 & 0.879 & 0.879 & 0.834 & 0.834 & 0.569 & 0.883 & +1.6 & 0.112 & 0.024 & 0.000 & 0.058 \\
ScienceQA & 4B & 2017 & 2--5 & 0.925 & 0.930 & 0.929 & 0.909 & 0.909 & 0.712 & 0.935 & +1.0 & 0.059 & 0.018 & 0.001 & 0.033 \\
ScienceQA & 8B & 2017 & 2--5 & 0.946 & 0.956 & 0.956 & 0.933 & 0.933 & 0.786 & 0.951 & +0.5 & 0.049 & 0.013 & 0.001 & 0.027 \\
ScienceQA & 7B & 2017 & 2--5 & 0.877 & 0.882 & 0.883 & 0.874 & 0.873 & 0.726 & 0.883 & +0.6 & 0.090 & 0.020 & 0.003 & 0.041 \\
CV-Bench & 2B & 2638 & 2--6 & 0.797 & 0.802 & 0.803 & 0.780 & 0.779 & 0.697 & 0.806 & +0.9 & 0.127 & 0.019 & 0.002 & 0.062 \\
CV-Bench & 4B & 2638 & 2--6 & 0.850 & 0.854 & 0.854 & 0.835 & 0.835 & 0.838 & 0.853 & +0.3 & 0.063 & 0.014 & 0.000 & 0.034 \\
CV-Bench & 8B & 2638 & 2--6 & 0.856 & 0.857 & 0.857 & 0.847 & 0.847 & 0.852 & 0.857 & +0.2 & 0.053 & 0.012 & 0.000 & 0.028 \\
CV-Bench & 7B & 2638 & 2--6 & 0.786 & 0.792 & 0.792 & 0.763 & 0.762 & 0.697 & 0.792 & +0.6 & 0.073 & 0.018 & 0.006 & 0.037 \\
A-OKVQA $K{=}8$ & 2B & 300 & 8 & 0.810 & 0.833 & 0.833 & 0.807 & 0.807 & 0.617 & 0.830 & +2.0 & 0.090 & 0.037 & 0.000 & 0.054 \\
A-OKVQA $K{=}8$ & 4B & 300 & 8 & 0.857 & 0.860 & 0.860 & 0.840 & 0.840 & 0.740 & 0.860 & +0.3 & 0.073 & 0.040 & 0.000 & 0.048 \\
A-OKVQA $K{=}8$ & 8B & 300 & 8 & 0.880 & 0.880 & 0.880 & 0.853 & 0.853 & 0.780 & 0.887 & +0.7 & 0.080 & 0.040 & 0.000 & 0.047 \\
A-OKVQA $K{=}8$ & 7B & 300 & 8 & 0.867 & 0.870 & 0.870 & 0.827 & 0.827 & 0.690 & 0.873 & +0.7 & 0.070 & 0.037 & 0.000 & 0.046 \\
A-OKVQA $K{=}16$ & 2B & 300 & 16 & 0.817 & 0.817 & 0.817 & 0.770 & 0.770 & 0.577 & 0.830 & +1.3 & 0.130 & 0.040 & 0.000 & 0.091 \\
A-OKVQA $K{=}16$ & 4B & 300 & 16 & 0.817 & 0.847 & 0.847 & 0.820 & 0.820 & 0.700 & 0.840 & +2.3 & 0.113 & 0.033 & 0.000 & 0.058 \\
A-OKVQA $K{=}16$ & 8B & 300 & 16 & 0.850 & 0.867 & 0.867 & 0.830 & 0.830 & 0.763 & 0.877 & +2.7 & 0.103 & 0.027 & 0.000 & 0.075 \\
A-OKVQA $K{=}16$ & 7B & 300 & 16 & 0.830 & 0.860 & 0.860 & 0.803 & 0.803 & 0.630 & 0.853 & +2.3 & 0.107 & 0.043 & 0.000 & 0.057 \\
EuroSAT & 2B & 1000 & 10 & 0.521 & 0.535 & 0.562 & 0.524 & 0.575 & 0.629 & 0.524 & +0.3 & 0.360 & 0.291 & 0.000 & 0.178 \\
EuroSAT & 4B & 1000 & 10 & 0.573 & 0.545 & 0.557 & 0.557 & 0.564 & 0.522 & 0.555 & -1.8 & 0.312 & 0.068 & 0.000 & 0.136 \\
EuroSAT & 8B & 1000 & 10 & 0.540 & 0.546 & 0.559 & 0.538 & 0.536 & 0.508 & 0.547 & +0.7 & 0.264 & 0.127 & 0.000 & 0.147 \\
EuroSAT & 7B & 1000 & 10 & 0.379 & 0.366 & 0.411 & 0.355 & 0.389 & 0.331 & 0.382 & +0.3 & 0.221 & 0.121 & 0.001 & 0.134 \\
Pets & 2B & 1000 & 26 & 0.757 & 0.808 & 0.830 & 0.818 & 0.849 & 0.835 & 0.808 & +5.1 & 0.259 & 0.106 & 0.000 & 0.130 \\
Pets & 4B & 1000 & 26 & 0.890 & 0.893 & 0.901 & 0.900 & 0.910 & 0.910 & 0.893 & +0.3 & 0.092 & 0.064 & 0.000 & 0.047 \\
Pets & 8B & 1000 & 26 & 0.876 & 0.905 & 0.908 & 0.899 & 0.910 & 0.901 & 0.905 & +2.9 & 0.124 & 0.076 & 0.000 & 0.058 \\
Pets & 7B & 1000 & 26 & 0.921 & 0.931 & 0.939 & 0.925 & 0.935 & 0.924 & 0.931 & +1.0 & 0.035 & 0.033 & 0.000 & 0.029 \\
MMMU & 2B & 845 & 2--9 & 0.479 & 0.489 & 0.489 & 0.417 & 0.417 & 0.334 & 0.497 & +1.8 & 0.453 & 0.201 & 0.000 & 0.218 \\
MMMU & 4B & 845 & 2--9 & 0.549 & 0.574 & 0.574 & 0.498 & 0.498 & 0.391 & 0.581 & +3.2 & 0.335 & 0.156 & 0.001 & 0.199 \\
MMMU & 8B & 845 & 2--9 & 0.560 & 0.574 & 0.574 & 0.536 & 0.536 & 0.444 & 0.583 & +2.4 & 0.325 & 0.140 & 0.000 & 0.198 \\
MMMU & 7B & 845 & 2--9 & 0.522 & 0.531 & 0.531 & 0.485 & 0.485 & 0.400 & 0.541 & +1.9 & 0.305 & 0.150 & 0.009 & 0.113 \\
\bottomrule
\end{tabular}
}
\caption{Comprehensive per-cell table for four Qwen models. A-OKVQA $K{=}8$ and $K{=}16$ are the synthetic expansions. L0+b and pw+b add the per-question prior; grey is pointwise plus the content-free prior at $\alpha{=}1$. Canonical columns rotate the text-sorted listing; cTV is the canonical correction TV. Flip raw and flip L0 are the caller-order reversal flips; flip L0$_{\mathrm{can}}$ is the canonical L0 flip. 
}
\label{tab:grid}
\end{table*}

\subsection{Position bias is real, and L0 usually helps}
\label{sec:position}
Table~\ref{tab:grid} gives the full per-cell caller and canonical grids, and more details are in Table \ref{tab:app-stage2} and Table \ref{tab:app-canonical}. Reversing the option list changes 3.5--36\% of answers on the non-MMMU datasets (see the column `flip raw'). The low end is Pets 7B
(3.5\%); the high end is EuroSAT 2B (36\%), where a
fixed class list gives ten positions to prefer. MMMU spans 30--45\%.

L0 cuts the flip by 1.1--6.6$\times$ (EuroSAT 2B 0.36$\to$0.29; CV-Bench 2B
0.13$\to$0.02) and improves the point estimate in 24 of the 28  cells
(2 ties, 2 losses; per-cell grid: Table~\ref{tab:grid}; intervals: Table~\ref{tab:app-interval}); under item-level bootstrap nine cells are nominally significant (no
multiplicity correction) and the only nominal loss is EuroSAT 4B. EuroSAT 4B drops from
0.573 to 0.545, a paired difference of $-2.8$ points with 95\% CI
$[-4.7,-1.0]$; 7B drops $-1.3$ points, CI $[-2.8,0.2]$, which is not
significant. 
Section~\ref{sec:confound} identifies the mechanism (fixed options); class
clustering widens the interval to include zero, so the mechanism is established
but the accuracy sign is not.


\subsection{Sorting removes listing ambiguity, not slot bias}
\label{sec:canonical}
Table~\ref{tab:grid} reports the canonical columns. With canonical order, raw keeps the caller listing and is unchanged (100\%
item agreement with the unsorted grid) and the sorted L0 reversal flip is zero in most cells, at most
0.95\%. The sort is input-order invariant: distinct texts give the same L0 prompts and
aligned distributions. The residual
45/32,400 item-reads (0.14\%) are exact probability ties: the argmax breaks ties
by slot index in the caller's array, naming a different option under reversal.
This is a decision tie-break, not a different sorted listing; option-text
tie-breaking gives zero. Appendix Table~\ref{tab:app-canonical} reports the full canonical grid.

It is not position-neutral. The single canonical read and its cycle average
differ by mean TV 0.027--0.22 across the  cells (MMMU 0.11--0.22). A
direct probe on A-OKVQA with 4B keeps the option set fixed and rotates the
canonical listing so every option visits every slot: the aligned distributions
differ by mean TV 0.0595 [0.0376, 0.0827], and accuracy over the four
rotations spans 0.855--0.875. Sorting removes the caller's ability to choose the slots; it does not
make the model indifferent to them.

\begin{table*}[t]
\centering
\small
\setlength{\tabcolsep}{4pt}
\begin{tabular}{@{}llrrrrrr@{}}
\toprule
Family & Dataset & raw & L0 & $\Delta$ [95\% CI] & flip raw & flip L0 & pointwise \\
\midrule
LLaVA-OneVision-7B & A-OKVQA & 0.913 & 0.923 & +1.0 [-0.333, +2.342] & 0.030 & 0.007 & 0.887 \\
LLaVA-OneVision-7B & ScienceQA & 0.963 & 0.960 & -0.3 [-1.667, +0.667] & 0.017 & 0.007 & 0.957 \\
LLaVA-OneVision-7B & EuroSAT & 0.497 & 0.453 & -4.3 [-9.667, +1.000] & 0.620 & 0.097 & 0.527 \\
LLaVA-OneVision-7B & Pets & 0.710 & 0.710 & +0.0 [-2.667, +2.667] & 0.153 & 0.070 & 0.650 \\
LLaVA-NeXT-7B & A-OKVQA & 0.843 & 0.857 & +1.3 [-0.333, +3.333] & 0.080 & 0.043 & 0.803 \\
LLaVA-NeXT-7B & EuroSAT & 0.497 & 0.503 & +0.7 [-4.333, +5.667] & 0.623 & 0.303 & 0.543 \\
LLaVA-NeXT-7B & Pets & 0.430 & 0.465 & +3.5 [-1.000, +8.500] & 0.540 & 0.425 & 0.535 \\
Idefics3-8B & A-OKVQA & 0.860 & 0.857 & -0.3 [-3.000, +2.333] & 0.120 & 0.047 & 0.860 \\
Idefics3-8B & EuroSAT & 0.347 & 0.380 & +3.3 [-0.667, +7.333] & 0.310 & 0.140 & 0.327 \\
Idefics3-8B & Pets & 0.590 & 0.673 & +8.3 [+4.000, +13.000] & 0.417 & 0.070 & 0.750 \\
\bottomrule
\end{tabular}
\caption{Cross-family reduced grids (300 items per cell; LLaVA-NeXT Pets 200). The pointwise column is the base readout.  }
\label{tab:app-cross}
\end{table*}

\subsection{The label prior works only where it is estimable}

\begin{table}[t]
\centering
\small
\setlength{\tabcolsep}{3pt}
\footnotesize
\setlength{\tabcolsep}{1.5pt}
\begin{tabular}{@{}lcccc@{}}
\toprule
 & \multicolumn{2}{c}{Pets} & \multicolumn{2}{c}{EuroSAT} \\
\cmidrule(lr){2-3}\cmidrule(lr){4-5}
Model & L0 & pointwise & L0 & pointwise \\
\midrule
2B & +2.2 [1.2, 3.2] & +3.1 [2.0, 4.3] & +2.7 [1.3, 4.2] & +5.1 [3.3, 7.1] \\
4B & +0.8 [0.2, 1.5] & +1.0 [0.4, 1.7] & +1.2 [0.0, 2.4] & +0.7 [-0.3, 1.7] \\
8B & +0.3 [-0.1, 0.8] & +1.1 [0.4, 1.9] & +1.3 [0.4, 2.3] & -0.2 [-1.2, 0.8] \\
7B & +0.8 [0.1, 1.6] & +1.0 [0.3, 1.8] & +4.5 [2.6, 6.4] & +3.4 [1.5, 5.3] \\
\bottomrule
\end{tabular}
\caption{Accuracy gain (pp) from adding the per-question batch prior to L0 or
to pointwise, with 95\% paired-bootstrap CIs. These are the only datasets on
which the prior is estimable. On a held-out-prefix split, gains over no prior
are Pets +1.8/+0.6/+0.2/+0.6 and EuroSAT +3.2/+1.2/+1.2/+4.0 for
2B/4B/8B/7B.}
\label{tab:prior}
\end{table}

\begin{figure*}[t]
\centering
\includegraphics[width=0.8\textwidth]{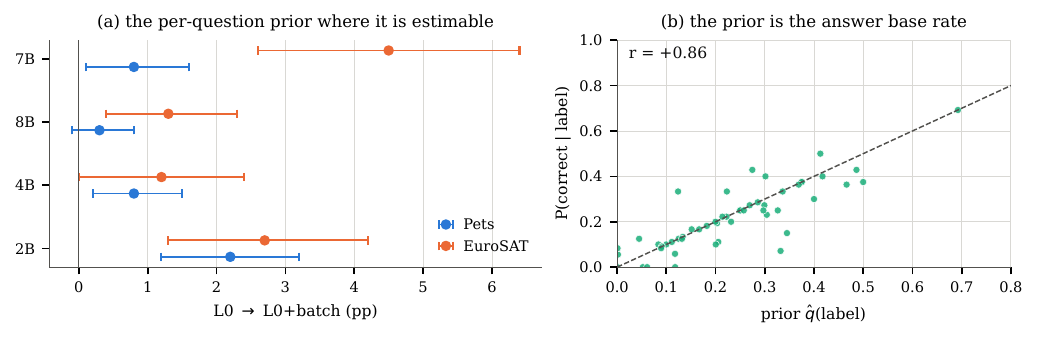}
\caption{(a) Adding the per-question prior to L0, where it is estimable.
(b) On open-set A-OKVQA the pooled prior tracks label correctness: the model's
mean probability for a label is close to the rate at which that label is
correct (Qwen3-VL-4B shown; the text gives the four-model range). Dividing it
removes signal, not bias.}
\label{fig:prior}
\end{figure*}

Table~\ref{tab:prior} reports the per-question prior gains, and Figure~\ref{fig:prior}a visualizes them. The prior needs at least eight states with the same question and
option list. Per-item VQA supplies almost none: zero on A-OKVQA, $K{=}8/16$ and MMMU;
65/2,017 on ScienceQA and 44/2,638 on CV-Bench ($\le0.5$ points). It activates on 993/1,000 fixed-question items and helps: Pets gains 0.3--2.2 points over L0 and EuroSAT 1.2--4.5 points
(Table~\ref{tab:prior}). On fixed-question sets this is a test-stream class prior (current item
included); a held-out prefix retains +0.2--+4.0 points (Table~\ref{tab:app-heldout}),
so part of the gain may be imbalance correction. Adding the same prior to
pointwise is the strongest configuration on Pets, 0.901 versus 0.884 for L0 and
0.885 for pointwise alone.

On open-set VQA the pooled substitute is not a bias estimate. Over the 54
A-OKVQA labels seen at least eight times, the prior correlates with correctness
at $+0.832$/$+0.864$/$+0.847$/$+0.855$ (2B/4B/8B/7B;
Figure~\ref{fig:prior}b); dividing it changes net correct by
$-7/-2/-2/-15$. The prior is the
answer base rate: signal.

\begin{table}[t]
\centering
\footnotesize
\setlength{\tabcolsep}{4pt}
\begin{tabular}{@{}lrrrr@{}}
\toprule
Family & Cells & Mean $\Delta$ & Sig.\ gains & Sig.\ losses \\
\midrule
Qwen grid (4 models) & 28 & +1.02 & 8 & 1 \\
LLaVA-OneVision-7B & 4 & +0.11 & 1 & 0 \\
LLaVA-NeXT-7B & 4 & +1.97 & 2 & 0 \\
Idefics3-8B & 4 & +3.37 & 3 & 0 \\
\bottomrule
\end{tabular}
\caption{Full-cell L0 accuracy change by family. Mean $\Delta$ is unweighted
over cells in pp; significant counts use item-bootstrap intervals excluding
zero. The Qwen row is the 28-cell grid; the non-Qwen rows are 12 full
cells spanning A-OKVQA, ScienceQA, EuroSAT and Pets. Per-cell values:
Table~\ref{tab:app-full}.  
}
\label{tab:crossfamily}
\end{table}

\subsection{Pointwise scoring usually loses}
 Table ~\ref{tab:grid} and   Table \ref{tab:app-methods}  give results for the pointwise method. On the Qwen grid, pointwise is 1.4--4.2 points worse than L0 on the non-Pets
VQA sets and 5.8 on MMMU; the gap grows with $K$: asking each option alone forgoes the contrast set
that the comparative readout uses, and more options make that set more
informative (the prompt also changes; Table~\ref{tab:app-context}). On this grid, averaged over models, pointwise beats L0 only on Pets, by
0.12 points (0.885 versus 0.884), and reaches 0.901 with the batch prior;
EuroSAT 4B is the per-cell exception (+1.2). Its listing invariance is exact for the batch-1 models (max
$|\Delta\pi|=2.2\times10^{-16}$); at 2B, batch 4, reversing the option order
changes $\pi$ by up to $2.6\times10^{-2}$ on Pets because the forward-batch
composition changes, a numerical rather than semantic order dependence.

\subsection{  Cross-family evaluation}
\label{sec:crossfamily}

Table \ref{tab:app-cross} shows the results for more models.
Reduced 300-item cells cut the reversal flip
(LLaVA-OneVision 0.030/0.017/0.620/0.153 $\to$ 0.007/0.007/0.097/0.070;
LLaVA-NeXT 0.080$\to$0.043 on A-OKVQA), but their accuracy deltas are mostly
non-significant; they bridge the Qwen grid and the full cells. Table~\ref{tab:crossfamily}
summarizes the full-cell result by family. The full cells change the picture:
LLaVA-NeXT gains $+2.2$ $[+0.9,+3.6]$ on A-OKVQA and $+4.6$ $[+2.5,+6.8]$ on
Pets; Idefics gains $+0.8$ ScienceQA, $+4.1$ EuroSAT and $+7.5$ Pets;
LLaVA-OneVision is $+0.7$ A-OKVQA, $+0.2$ ScienceQA, $+1.6$ Pets and $-2.1$
EuroSAT. L0 reduces the reversal flip in all 12 full cells; the accuracy sign, not stability, is
what varies. A 40-cell family model gives accuracy $+1.44$ $[+0.60,+2.28]$: gains are
family- and dataset-dependent. The effect is carried by the non-Qwen full cells;
the Qwen-only aggregate remains $+0.72$ $[-0.39,+1.83]$. Across cells, family
ranges are Qwen $-2.8$--$5.1$, LLaVA-OneVision $-2.1$--$1.6$, LLaVA-NeXT
$-0.1$--$4.6$, and Idefics3 $0.8$--$7.5$ points. Full-cell pointwise is
family-dependent (Table~\ref{tab:app-pointwise-full}): LLaVA-OneVision wins
EuroSAT ($+5.5$) but loses Pets ($-3.7$); NeXT wins EuroSAT ($+3.2$) and Pets
($+0.3$); Idefics3 wins Pets ($+5.7$) but loses EuroSAT ($-7.4$). The batch
prior makes pointwise best on Pets for all three ($+11.5/+8.9/+8.6$).
LLaVA-NeXT replicates the full protocol; Tables~\ref{tab:app-next-canonical}
and~\ref{tab:app-next-protocol} give its canonical, prior, grey, and context
results. The options-context control is mixed: on NeXT it lowers pointwise
accuracy in every cell, while on Qwen 2B/4B the full-A-OKVQA change is
$-2.1/+0.7$ and EuroSAT $+0.5/-5.0$ (Table~\ref{tab:app-context}). Showing the
alternatives therefore does not isolate the contrast set as a single factor.

\subsection{A content-free prior works only when it is aligned}
\label{sec:grey}
\begin{figure}[t]
\centering
\includegraphics[width=1\columnwidth]{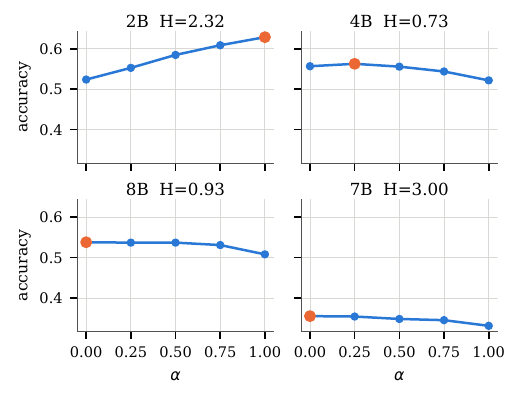}
\caption{EuroSAT pointwise accuracy as the grey prior strength $\alpha$ grows.
The orange dot is the test-set best $\alpha$ and is optimistic; the held-out
estimate is in Section~\ref{sec:grey}. Only 2B, whose grey and real marginals both
over-predict brushland, improves monotonically.}
\label{fig:grey}
\end{figure}

Figure~\ref{fig:grey} reports the $\alpha$ sweep. On EuroSAT 2B, $\alpha{=}1$ raises pointwise accuracy from 0.524 to 0.629
(+10.5 points, CI [7.1, 13.9]); $\alpha{=}0.5$ gives +6.1 points and improves
ECE from 0.152 to 0.042 (Figure~\ref{fig:grey}). The grey marginal over-predicts
brushland (0.501) and the real marginal over-predicts it too (0.258), so
subtracting the grey log-odds removes a genuine content-free bias. The same
$\alpha{=}1$ correction hurts EuroSAT 4B ($-3.5$, CI $[-5.8,-1.1]$), 8B
($-3.0$) and 7B ($-2.4$): 4B's concentrated grey marginal (entropy 0.73 bits)
overcorrects, and 8B's points at lake/sea (0.804) while the real errors are
brushland/residential.
Table~\ref{tab:app-grey-diagnostic} makes the alignment condition concrete.
The two cells selected by matching grey/real top-1 classes and grey entropy above
1.5 bits are EuroSAT 2B ($+10.5$) and Pets 2B ($+1.7$); other positive cells
gain at most $1.0$. EuroSAT 4B matches top-1 but has entropy 0.73 and overcorrects;
mismatched cells are small or harmful. This is a diagnostic, not a universal rule.
On A-OKVQA it loses 4.2--8.6 points for every model: grey-only accuracy is
0.41--0.45, so the content-free prompt still carries answer-frequency
information. 

As shown in Table \ref{tab:grid}, with $\alpha$ fixed at 1, the grey prior is neutral on average on Pets (0.893
versus 0.885 for pointwise alone) and EuroSAT (0.498 versus 0.493), with
cell-level gains from +10.5 to losses of $-3.5$, and harmful elsewhere; selecting $\alpha$ on half of each cell and
evaluating on the other half gives 0.517 vs 0.508 on EuroSAT and 0.887 vs 0.882
on Pets, with no gain elsewhere. 

Appendix \ref{sec:app_ablations} discusses more results on additional ablations, e.g.,   prompt formats move accuracy much less, but option wording matters more.

 \section{When Does Debiasing Transfer?}
\label{sec:analysis}

\subsection{The fixed-option confound}
\label{sec:confound}

EuroSAT and Pets show the same class list for every image, so a class is always
at the same slot: annual crop at slot 0, pasture at slot 5, and so on. The raw
read's position preference is therefore also a class preference, and L0 cannot
remove one without the other. On EuroSAT 4B the raw position marginal
over-predicts early slots (annual crop 0.176, forest 0.227; the true shares are
about 0.10) and suppresses pasture at slot 5 (0.031 versus a true 0.067). L0
removes that advantage. Table~\ref{tab:eurosat} shows the consequence: pasture
improves from 0.30 to 0.51 and river from 0.62 to 0.66, while annual crop falls
from 0.55 to 0.36 and brushland from 0.46 to 0.32. The correlation between the
raw class marginal and (raw accuracy $-$ L0 accuracy) is $+0.64$ at 4B, $+0.51$
at 2B, $+0.47$ at 7B and $+0.03$ at 8B. The sign is model-dependent in the point estimates: L0 loses at 4B and 7B but helps
at 2B/8B. Slot randomization reverses the 2B/4B/7B signs
($+1.7\to-0.3$, $-1.0\to+2.0$, $-1.0\to+1.0$), while 8B stays $+1.7$ and Pets
2B stays $+4.7\to+3.7$ (300-item; Tables~\ref{tab:app-slot},~\ref{tab:app-slot-full}).

\begin{table}[t]
\centering
\small
\setlength{\tabcolsep}{3.5pt}
\begin{tabular}{@{}lrrrrr@{}}
\toprule
Class & $n$ & raw & L0 & G$-$R & texture \\
\midrule
annual crop & 107 & 0.55 & 0.36 & $-0.04$ & 0.101 \\
brushland & 115 & 0.46 & 0.32 & $0.00$ & 0.061 \\
forest & 120 & 0.82 & 0.74 & $+0.10$ & 0.013 \\
pasture & 67 & 0.30 & 0.51 & $+0.10$ & 0.038 \\
river & 85 & 0.62 & 0.66 & $+0.07$ & 0.075 \\
permanent crop & 103 & 0.00 & 0.04 & $-0.03$ & 0.083 \\
\bottomrule
\end{tabular}
\caption{EuroSAT 4B by class: raw and L0 accuracy, colour (G$-$R) and texture
(gray standard deviation). Pasture and river gain under L0; annual crop,
brushland and forest lose. The fixed class order ties each class to a slot.}
\label{tab:eurosat}
\end{table}

\subsection{The label prior must be identifiable}

A prior is useful only if it estimates a bias rather than the answer
distribution. In per-item VQA each question is asked once, so the per-question
prior is the identity. The pooled substitute is the model's answer marginal: on
A-OKVQA it correlates $+0.83$ to $+0.86$ with correctness, and dividing it costs
up to 15 items. Compute this correlation on a labelled development set: near
zero it is a bias candidate; clearly positive it encodes correctness and should
be left alone. On fixed-question data the same diagnostic applies per class.

\subsection{Pointwise forgoes the contrast set}

On the Qwen grid, pointwise is order-invariant by construction, but it forgoes
the list of alternatives. On EuroSAT 4B it collapses on heterogeneous classes
(annual crop 0.55 $\to$ 0.23, forest 0.82 $\to$ 0.56); its top-1/top-2 margin
is smaller (0.760 versus 0.849), and it changes 28\% of answers relative to raw.
Pets is the exception: 26 distinctive breeds, 0.88--0.92 raw accuracy at
4B/8B/7B, and a breed marginal close to the truth (entropy 4.646 versus 4.69 at
4B). Option identity is nearly unambiguous there, so pointwise is competitive
and becomes best only after the batch prior.

\subsection{A diagnostic checklist}

Run these checks before choosing a correction; each produces a diagnostic to
report, not a binary pass/fail:
\begin{enumerate}
\item \textbf{Order audit.} Compare the caller listing with its reversal on a
labelled audit set. Our reduced grids use 200--300 items and the full grids
use more; 100--200 items is a practical minimum. Report the argmax flip,
aligned reversal TV, and correction TV between one canonical read and its
cycle average. A small flip with a large correction TV means slot bias
remains after sorting.
\item \textbf{Prior estimability.} Count the states sharing each
question-and-option-list key. Fewer than eight states make the per-question
prior the identity; a pooled prior then estimates the answer base rate, not
bias. Across option labels, correlate the candidate prior with empirical
$P(\text{correct}\mid\text{label})$ on a labelled development set (at least
50--100 items; our A-OKVQA diagnostic uses all 1,145 items and 54 option
labels with at least eight occurrences). Near zero supports division; above about 0.5 the prior
carries correctness signal and should not be divided.
\item \textbf{Fixed-list confound.} Randomize the class-to-slot assignment and
repeat the L0 comparison. If the accuracy sign flips relative to the original
assignment, position is aliased with class; also report the correlation
between the raw class marginal and raw$-$L0 accuracy
(Section~\ref{sec:confound}).
\item \textbf{Readout comparison.} On one held-out set, compare raw, L0,
L0+prior, pointwise and pointwise+prior. Report accuracy with paired bootstrap
intervals, reversal flip/TV, ECE, and coverage at 5\% risk; use item and class
resampling for fixed-option sets.
\item \textbf{Grey prior.} Use only for fixed-option classification. Compute
the mean pointwise marginal on uniform-grey images: its entropy, top-1 class,
and overlap with the real marginal. Apply a correction only when top-1 classes
match and grey entropy is above about 1.5 bits; a concentrated grey marginal
can overcorrect even when top-1 matches (EuroSAT 4B, $H{=}0.73$;
Table~\ref{tab:app-grey-diagnostic}). Select $\alpha$ on a held-out half and
treat fewer than 50 items as insufficient.
\end{enumerate}
Steps 3 and 5 apply only to fixed-option classification; steps 1, 2 and 4
apply to both settings. Steps 1--2 need roughly 100--200 labelled items. All
thresholds are conservative heuristics, not calibrated operating points;
report the observed margins.

\section{Conclusion}
\label{sec:conclusion}

Training-free debiasing transfers only when bias is identifiable: repeated
questions, free options, informative contrast. Stability transfers; accuracy
is family-dependent.

{
    \small
    \bibliographystyle{ieeenat_fullname}
    \bibliography{main}
}

\clearpage
\appendix
\setcounter{table}{0}
\setcounter{figure}{0}
\renewcommand{\thetable}{A\arabic{table}}
\renewcommand{\thefigure}{A\arabic{figure}}

\section{Experiment Details}
\label{sec:appendix}

This appendix gives the construction, prompts, hyperparameters, full per-cell
numbers and reproduction commands behind the main paper.   The dataset and model
tables summarize the loaders and run metadata.

\begin{table}[t]
\centering
\small
\setlength{\tabcolsep}{3pt}
\begin{tabular}{@{}lrrp{2.7cm}@{}}
\toprule
Dataset & $n$ & $K$ & Character \\
\midrule
A-OKVQA (full) & 1,145 & 3--4 & open-set VQA reference \\
ScienceQA & 2,017 & 2--5 & per-item, hint text \\
CV-Bench & 2,638 & 2--6 & per-item \\
A-OKVQA $K{=}8$ & 300 & 8 & synthetic distractors \\
A-OKVQA $K{=}16$ & 300 & 16 & synthetic distractors \\
EuroSAT & 1,000 & 10 & fixed classes, 64$\times$64 \\
Pets & 1,000 & 26 & fixed breeds \\
MMMU & 845 & 2--9 & multi-image \\
\bottomrule
\end{tabular}
\caption{Evaluation sets. $K$ is the number of options. The 26-option cap is
the letter readout's A--Z limit.}
\label{tab:data}
\end{table}

\begin{table}[t]
\centering
 \small
\setlength{\tabcolsep}{4pt}
\begin{tabular}{@{}lrrr@{}}
\toprule
Model & batch & full-list fwd & option fwd \\
\midrule
2B & 4 & 126.6 & 127.0 \\
4B & 1 & 56.5 & 54.5 \\
8B & 1 & 87.1 & 83.4 \\
7B & 1 & 103.9 & 101.6 \\
\bottomrule
\end{tabular}
\caption{Batch-adjusted mean latency (ms) of one forward over the seven
 sets. A full-list forward is one L0 shift; an option forward is one
pointwise probe. Per-cell ranges are 30--271 ms and 27--273 ms, driven by image
size and prompt length.  
}
\label{tab:runtime}
\end{table}

\subsection{Datasets and preprocessing}

\begin{table*}[t]
\centering
\footnotesize
\setlength{\tabcolsep}{4pt}
\begin{tabular}{@{}lp{2.7cm}rrp{2.8cm}p{2.5cm}@{}}
\toprule
Set & Source / split & $n$ & $K$ & Options & License \\
\midrule
A-OKVQA & HF validation & 1,145 & 3--4 & per-item & Apache-2.0$^\dagger$ \\
ScienceQA & test, image items & 2,017 & 2--5 & per-item + hint & CC-BY-SA-4.0 \\
CV-Bench & test & 2,638 & 2--6 & per-item & Apache-2.0 \\
A-OKVQA $K{=}8$ & synthetic & 300 & 8 & fixed per item & Apache-2.0$^\dagger$ \\
A-OKVQA $K{=}16$ & synthetic & 300 & 16 & fixed per item & Apache-2.0$^\dagger$ \\
EuroSAT & test subset & 1,000 & 10 & fixed 10 classes & MIT \\
Pets & test subset & 1,000 & 26 & fixed 26 breeds & CC-BY-SA-4.0 \\
MMMU & validation, MC & 845 & 2--9 & per-item, multi-image & Apache-2.0 \\
\bottomrule
\end{tabular}
\caption{Evaluation sets. EuroSAT and Pets use seeded 1,000-item test subsets;
matched EuroSAT cells use the same 1,000-item subset; the Idefics3 full-split
2,700-item run is a robustness check. $^\dagger$The A-OKVQA repository is Apache-2.0; its images are COCO
images under COCO's terms.}
\label{tab:app-data}
\end{table*}

Table~\ref{tab:app-data} lists the evaluation sets. A-OKVQA uses the validation split and the supplied choices; the full set has
1,138 four-option and 7 three-option items. The synthetic $K{=}8/16$ sets start
from the same 300 validation items, append unique distractors sampled from the
global A-OKVQA choice pool, and shuffle the final list; this is a controlled
large-$K$ stress test, not a natural large-$K$ benchmark. ScienceQA keeps only
test items with an image and prepends the hint text when present. MMMU keeps
every multiple-choice validation item with at least one image and at most 26
options. Its \texttt{options} field is sometimes a string such as
\texttt{"['\$6', '\$7']"}; the adapter parses it with
\texttt{ast.literal\_eval} before deduplication. Runs made before that fix split
the string into characters and are quarantined. The main
grid uses a seeded 1,000-item subset of EuroSAT's 2,700-image test split for
matched cross-family cells; the Idefics3 full-split 2,700 run is a robustness
check. Pets keeps the 26 largest test classes
because the letter readout stops at Z and samples 1,000 of their images. Option
strings are deduplicated in every adapter. Dataset selection uses seed 0; the
synthetic distractors use seed $i$ for item $i$.

\subsection{Models and environment}

\begin{table}[t]
\centering
\footnotesize
\setlength{\tabcolsep}{3pt}
\begin{tabular}{@{}lrrl@{}}
\toprule
Model & Params & Batch & Role \\
\midrule
Qwen3-VL-2B-Instruct & 2B & 4 & main grid \\
Qwen3-VL-4B-Instruct & 4B & 1 & main grid \\
Qwen3-VL-8B-Instruct & 8B & 1 & main grid \\
Qwen2.5-VL-7B-Instruct & 7B & 1 & main grid \\
LLaVA-OneVision-7B & 7B & 1 & cross-family \\
LLaVA-NeXT-7B & 7B & 1 & cross-family \\
Idefics3-8B & 8B & 1 & cross-family \\
\bottomrule
\end{tabular}
\caption{Models. All runs use bfloat16, a $448^2$-pixel image budget and
\texttt{model.eval()} with no gradient.}
\label{tab:app-models}
\end{table}

Table~\ref{tab:app-models} lists the models. All runs are on one NVIDIA RTX 4090 D (24 GB), driver 595.99.02, on an
i9-14900K. Four faulty logical CPUs (8--11) are offlined; the affected cells
were rerun with CPU affinity and are otherwise identical. Software versions are
Python 3.12.4, torch 2.14.1+cu126, transformers 5.20.2 and anyjev 0.3.0. Each
per-cell JSON records these values, the date, GPU, batch size and dtype; the
later full-cell runs also record the driver.

\subsection{Runtime.}\label{sec:app_runtime}
All timings are on one RTX 4090 D at bfloat16, batch 4 for 2B and batch 1 for
the other models (Table~\ref{tab:runtime}). A full-list L0 shift costs 127, 56,
87 and 104 ms on average for 2B/4B/8B/7B; a single-option pointwise forward
costs 127, 54, 83 and 102 ms. The option prompt is not materially cheaper
(0.96--1.00$\times$): the image prefix dominates. The per-item cost follows the
number of forwards: raw is one, L0 and pointwise are $K$, the grey pass is $K$
option forwards, and the prior study is $1+2K$ because it runs raw and both L0
variants. For the $K{=}10$ EuroSAT and $K{=}26$ Pets cells, measured L0 costs
0.11--0.43 / 0.51--2.00 s per item, pointwise 0.11--0.38 / 0.42--1.64 s, and
the prior study 0.25--0.91 / 1.03--4.11 s. The 96 prior/pointwise/grey jobs
behind the main grid took 20.4 GPU-hours: 8.0 h for A-OKVQA, Pets and EuroSAT
and 12.4 h for the five-dataset extension, including model loading and
checkpointing; the separate  raw/L0/reversal grid is not counted. The three non-Qwen
families were run separately.

\subsection{Prompts}

The readout uses the processor's chat template. The system prompt is:
\begin{quote}\ttfamily\small
You are a decision function. You will be given a state and one question.\\
Reply with the answer label only: no words, no punctuation, no explanation.
\end{quote}
For the raw and L0 readouts, the user content is:
\begin{quote}\ttfamily\small
State:\\
(empty)\\
\\
Question: question\\
Options:\\
A. option 1\\
B. option 2\\
\ldots\\
Answer with the letter only.
\end{quote}
L0 repeats this block under each cyclic shift of the option list; canonical L0
first sorts the options by text. Images replace the state markers; the system
prompt and the answer suffix are unchanged. The pointwise runner uses:
\begin{quote}\ttfamily\small
You are a decision function. You will be given a state, a question and one
option. Reply with yes or no only: no words, no punctuation, no explanation.
\end{quote}
\begin{quote}\ttfamily\small
State: image\\
\\
Question: question\\
Option: option\\
Is this option the correct answer? Answer yes or no.
\end{quote}
The grey probe is the pointwise prompt with every image replaced by a uniform
grey image of the same size. Labels are the letters A--Z; we require a
single-token label for both the bare and leading-space forms. Pointwise yes/no
tokens are collected over the same variants (lowercase, capitalised, leading
space), and the run aborts if yes and no share a token id.

\subsection{Readout and prior hyperparameters}

\begin{itemize}
\item \textbf{raw}: one prefill, caller listing.
\item \textbf{L0}: all $K$ cyclic shifts, log-space mean, no shift cap.
\item \textbf{canonical}: sort by \texttt{str(option)} with caller index as
tie-break; the loaders deduplicate, so the key is unique.
\item \textbf{batch prior}: minimum 8 states per question, strength
$\alpha{=}0.75$, applied to each shift before marginalization.
\item \textbf{pooled prior}: same strength, keyed by $K$ for L0 and by option
text for pointwise.
\item \textbf{pointwise}: yes/no log-odds over one option at a time; base
readout has no prior; the +batch and +pooled variants use the same
$\alpha{=}0.75$ and minimum 8 states.
\item \textbf{grey}: $\alpha\in\{0.25,0.5,0.75,1.0\}$ in the sweep; the main
table uses the package default $\alpha{=}1$. Held-out selection splits each
cell 50/50, chooses $\alpha$ on the first half and reports the second.
\item \textbf{bootstrap}: 2,000--4,000 percentile resamples. The main intervals
resample items; the fixed-option sets also resample true classes.
\end{itemize}

\paragraph{Code-level details.}
The readout never samples: \texttt{Decider} reads the final-position logits,
restricts them to the single-token label variants (bare, leading space,
capitalised), and normalizes. L0 builds the $K$ cyclic prompts, divides each
shift's label distribution by the prior, and averages in log space before the
final softmax. Canonical order sorts by \texttt{str(option)} with the
caller index as tie-break; deduplication makes the key unique. Pointwise builds
one prompt per option, sums the yes and no log-probabilities over the same token
variants, and softmaxes the log-odds over options. The vision backend renders
the processor's chat template; LLaVA-NeXT and Idefics3 use structured content
parts. Images use \texttt{max\_pixels}$=448^2$, bfloat16, and no gradients.
Bootstrap intervals are percentile resamples (2,000--4,000); fixed-option sets
resample true classes in addition to items.

\subsection{Released artifact and reproduction}
\label{sec:app-artifact}

The supplementary folder (\texttt{supplemental/}, also uploaded as
\texttt{supplemental.zip}) is an anonymous, Apache-2.0 artifact. It contains the
training-free readout and vision-language backend (\texttt{anyjev/}), the
dataset adapters and experiment drivers (\texttt{scripts/}), CPU tests
(\texttt{tests/}), and the summary and per-item JSONs behind the tables
(\texttt{research/visual\_bias/}). It contains no model weights and no dataset
images; the loaders download the public checkpoints and evaluation splits. The
uncompressed folder is below 200 MB. In particular, the Qwen grids are
represented by per-cell summaries, while the non-Qwen pointwise cells and the
ablation per-item records are retained.

\paragraph{Install and run.}
From the artifact root:
\begin{quote}\ttfamily\small
python -m pip install -e ".[hf,vlm,data,dev]"\\
python scripts/eval\_visual\_grid.py --dataset scienceqa \\
\quad --models Qwen/Qwen3-VL-4B-Instruct --limit 2017 --batch-size 1 \\
\quad --out-dir research/visual\_bias/stage2
\end{quote}
The pointwise, prior, grey, permutation, format, paraphrase and
slot-randomization drivers accept the same \texttt{--dataset}, \texttt{--models},
\texttt{--limit}, \texttt{--batch-size} and \texttt{--out-dir} arguments; the
grey driver also takes \texttt{--pointwise-dir} and \texttt{--alphas}. Their
names are \texttt{eval\_visual\_pointwise.py}, \texttt{eval\_visual\_priors.py},
\texttt{eval\_visual\_pointwise\_grey.py},
\texttt{eval\_visual\_permutations.py}, \texttt{eval\_visual\_formats.py},
\texttt{eval\_visual\_paraphrase.py} and
\texttt{eval\_visual\_slotrandom.py}; the text-domain control is
\texttt{eval\_text\_control.py}. Images are cached under
\texttt{research/visual\_bias/stage2\_images/} and datasets under
\texttt{/tmp/anyjev\_data} by default; both directories are configurable. The
model IDs and batch sizes are those in Table~\ref{tab:app-models}.

\paragraph{Data-to-table map.}
The main per-cell table (Table~\ref{tab:grid}) is rebuilt from
\texttt{stage2/overall\_grid\_summary.json},
\texttt{stage2\_canonical/overall\_grid\_summary.json} and
\texttt{all\_methods\_summary.json}. The reduced cross-family table
(Table~\ref{tab:app-cross}) uses \texttt{analysis/cross\_family.json} together
with the \texttt{nonqwen*/stage2} and \texttt{nonqwen*/pointwise} files. The
full non-Qwen cells (Table~\ref{tab:app-full}) use the matching
\texttt{nonqwen*\_full/stage2*} summaries and
\texttt{nonqwen*\_full/pointwise\_full/*.json}. The grey diagnostic
(Table~\ref{tab:app-grey-diagnostic}) uses
\texttt{pilot\_pointwise\_grey/}. The derived \texttt{analysis/} JSONs contain
the bootstrap intervals, family models, runtime table and held-out grey
selection; \texttt{pilot\_prior/activation\_counts.json} records the
per-question-prior activation counts extracted from the per-cell prior runs.

\paragraph{Validation.}
The CPU-only tests exercise visual-state rendering, dataset parsing and
deduplication, canonical ordering, pointwise scoring, priors and metrics with a
fake backend:
\begin{quote}\ttfamily\small
pytest -q tests/test\_visual*.py
\end{quote}
They report 51 passed and 1 skipped on CPU without a GPU. Real-engine smoke
tests are omitted because they require a GPU and a model download.

Dataset and model terms are summarized in
\texttt{THIRD\_PARTY.md}; they are downloaded at run time and are not
redistributed. The paper's claims concern the typed, label-restricted readout
implemented in \texttt{anyjev/}, not generative VLM answering.

\subsection{Full per-cell results}

Tables~\ref{tab:app-stage2}--\ref{tab:app-slot} give every cell behind the
main figures: the uncorrected grid, the canonical grid, the method comparison
(Table~\ref{tab:app-methods}), the cross-family evaluation
(Table~\ref{tab:app-cross}) and the slot-randomization ablation
(Table~\ref{tab:app-slot}).
Table~\ref{tab:app-full} adds full-scale cells for LLaVA-OneVision,
LLaVA-NeXT and Idefics3 across A-OKVQA, ScienceQA, EuroSAT and Pets; it
confirms the reduced-grid flip signs and shows where full-set accuracy gains
differ by family. The main text
reports means over models; the appendix is the place to see whether a mean is
carried by one model. In particular, the EuroSAT 4B loss and the Pets 2B gain
are visible per cell, and the canonical correction TV is separated from the
reversal flip. Table~\ref{tab:app-ece} reports ECE; L0 lowers it in 25 of 28
cells. The exceptions are CV-Bench 7B (+0.002), EuroSAT 4B (+0.025) and
EuroSAT 7B (+0.013). Table~\ref{tab:app-interval} gives per-cell bootstrap
intervals for the flip and TV estimates. Per-cell L0$-$raw ranges 0.17--1.14 on
full A-OKVQA, 0.3--5.1 on Pets and $-2.8$ to 1.4 on EuroSAT; 4B is the worst cell.
The Idefics3 EuroSAT row in Table~\ref{tab:app-full} is the matched 1,000-item
subset; its full 2,700-item run gives .347/.378 and flip .287$\to$.122.
Across 40 cells from four families (the 28-cell Qwen grid plus 12 full-scale
non-Qwen cells), a family-level mixed model gives accuracy $+1.44$
$[+0.60,+2.28]$; family means are Qwen 1.02, LLaVA-OneVision 0.11,
LLaVA-NeXT 1.97 and Idefics3 3.37 points.

\begin{table*}[t]
\centering
\scriptsize
\setlength{\tabcolsep}{3pt}
\begin{tabular}{@{}llrrrrrrrr@{}}
\toprule
Dataset & Model & $n$ & $K$ & raw & L0 & $\Delta$ [95\% CI] & flip raw & flip L0 & TV raw / L0 \\
\midrule
ScienceQA & 2B & 2017 & 2--5 & 0.867 & 0.879 & +1.14 [+0.10, +2.13] & 0.112 & 0.024 & 0.100 / 0.026 \\
ScienceQA & 4B & 2017 & 2--5 & 0.925 & 0.930 & +0.50 [-0.35, +1.34] & 0.059 & 0.018 & 0.057 / 0.018 \\
ScienceQA & 8B & 2017 & 2--5 & 0.946 & 0.956 & +1.04 [+0.40, +1.69] & 0.049 & 0.013 & 0.050 / 0.011 \\
ScienceQA & 7B & 2017 & 2--5 & 0.877 & 0.882 & +0.50 [-0.35, +1.34] & 0.090 & 0.020 & 0.072 / 0.016 \\
CV-Bench & 2B & 2638 & 2--6 & 0.797 & 0.802 & +0.53 [-0.42, +1.44] & 0.127 & 0.019 & 0.109 / 0.013 \\
CV-Bench & 4B & 2638 & 2--6 & 0.850 & 0.854 & +0.42 [-0.27, +1.02] & 0.063 & 0.014 & 0.059 / 0.012 \\
CV-Bench & 8B & 2638 & 2--6 & 0.856 & 0.857 & +0.19 [-0.38, +0.80] & 0.053 & 0.012 & 0.046 / 0.010 \\
CV-Bench & 7B & 2638 & 2--6 & 0.786 & 0.792 & +0.57 [-0.15, +1.33] & 0.073 & 0.018 & 0.068 / 0.007 \\
A-OKVQA $K{=}8$ & 2B & 300 & 8 & 0.810 & 0.833 & +2.33 [+0.67, +4.01] & 0.090 & 0.037 & 0.093 / 0.036 \\
A-OKVQA $K{=}8$ & 4B & 300 & 8 & 0.857 & 0.860 & +0.33 [-1.33, +2.00] & 0.073 & 0.040 & 0.070 / 0.035 \\
A-OKVQA $K{=}8$ & 8B & 300 & 8 & 0.880 & 0.880 & +0.00 [-2.33, +2.33] & 0.080 & 0.040 & 0.079 / 0.035 \\
A-OKVQA $K{=}8$ & 7B & 300 & 8 & 0.867 & 0.870 & +0.33 [-1.33, +2.00] & 0.070 & 0.037 & 0.080 / 0.033 \\
A-OKVQA $K{=}16$ & 2B & 300 & 16 & 0.817 & 0.817 & +0.00 [-3.33, +3.33] & 0.130 & 0.040 & 0.133 / 0.048 \\
A-OKVQA $K{=}16$ & 4B & 300 & 16 & 0.817 & 0.847 & +3.00 [+1.00, +5.34] & 0.113 & 0.033 & 0.099 / 0.039 \\
A-OKVQA $K{=}16$ & 8B & 300 & 16 & 0.850 & 0.867 & +1.67 [-0.33, +4.00] & 0.103 & 0.027 & 0.101 / 0.037 \\
A-OKVQA $K{=}16$ & 7B & 300 & 16 & 0.830 & 0.860 & +3.00 [+0.67, +5.33] & 0.107 & 0.043 & 0.099 / 0.038 \\
EuroSAT & 2B & 1000 & 10 & 0.521 & 0.535 & +1.40 [-0.60, +3.30] & 0.360 & 0.291 & 0.331 / 0.225 \\
EuroSAT & 4B & 1000 & 10 & 0.573 & 0.545 & -2.80 [-4.70, -1.00] & 0.312 & 0.068 & 0.300 / 0.076 \\
EuroSAT & 8B & 1000 & 10 & 0.540 & 0.546 & +0.60 [-1.40, +2.60] & 0.264 & 0.127 & 0.249 / 0.108 \\
EuroSAT & 7B & 1000 & 10 & 0.379 & 0.366 & -1.30 [-2.80, +0.30] & 0.221 & 0.121 & 0.195 / 0.111 \\
Pets & 2B & 1000 & 26 & 0.757 & 0.808 & +5.10 [+3.30, +7.00] & 0.259 & 0.106 & 0.252 / 0.101 \\
Pets & 4B & 1000 & 26 & 0.890 & 0.893 & +0.30 [-1.00, +1.60] & 0.092 & 0.064 & 0.092 / 0.065 \\
Pets & 8B & 1000 & 26 & 0.876 & 0.905 & +2.90 [+1.40, +4.40] & 0.124 & 0.076 & 0.124 / 0.076 \\
Pets & 7B & 1000 & 26 & 0.921 & 0.931 & +1.00 [+0.20, +1.80] & 0.035 & 0.033 & 0.053 / 0.036 \\
MMMU & 2B & 845 & 2--9 & 0.479 & 0.489 & +0.95 [-1.78, +3.79] & 0.453 & 0.201 & 0.349 / 0.124 \\
MMMU & 4B & 845 & 2--9 & 0.549 & 0.574 & +2.49 [-0.12, +5.09] & 0.335 & 0.156 & 0.310 / 0.113 \\
MMMU & 8B & 845 & 2--9 & 0.560 & 0.574 & +1.42 [-1.07, +3.91] & 0.325 & 0.140 & 0.293 / 0.113 \\
MMMU & 7B & 845 & 2--9 & 0.522 & 0.531 & +0.95 [-1.54, +3.31] & 0.305 & 0.150 & 0.183 / 0.057 \\
\bottomrule
\end{tabular}
\caption{Full  grid. $\Delta=100(\mathrm{L0}-\mathrm{raw})$ in percentage points; CI bounds are ordered and must contain $\Delta$ (asserted by the generator). TV is mean aligned total variation between original and reversed distributions. }
\label{tab:app-stage2}
\end{table*}

\begin{table*}[t]
\centering
\scriptsize
\setlength{\tabcolsep}{3pt}
\begin{tabular}{@{}llrrrrrr@{}}
\toprule
Dataset & Model & raw & L0 & $\Delta$ [95\% CI] & flip raw & flip L0 & correction TV \\
\midrule
ScienceQA & 2B & 0.867 & 0.883 & +1.59 [+0.545, +2.628] & 0.112 & 0.000 & 0.058 \\
ScienceQA & 4B & 0.925 & 0.935 & +0.99 [+0.198, +1.785] & 0.059 & 0.001 & 0.033 \\
ScienceQA & 8B & 0.946 & 0.951 & +0.50 [-0.198, +1.190] & 0.049 & 0.001 & 0.027 \\
ScienceQA & 7B & 0.877 & 0.883 & +0.64 [-0.198, +1.537] & 0.090 & 0.003 & 0.041 \\
CV-Bench & 2B & 0.797 & 0.806 & +0.87 [-0.076, +1.820] & 0.127 & 0.002 & 0.062 \\
CV-Bench & 4B & 0.850 & 0.853 & +0.27 [-0.417, +0.948] & 0.063 & 0.000 & 0.034 \\
CV-Bench & 8B & 0.856 & 0.857 & +0.15 [-0.455, +0.758] & 0.053 & 0.000 & 0.028 \\
CV-Bench & 7B & 0.786 & 0.792 & +0.61 [-0.152, +1.365] & 0.073 & 0.006 & 0.037 \\
A-OKVQA $K{=}8$ & 2B & 0.810 & 0.830 & +2.00 [+0.000, +4.000] & 0.090 & 0.000 & 0.054 \\
A-OKVQA $K{=}8$ & 4B & 0.857 & 0.860 & +0.33 [-2.000, +2.667] & 0.073 & 0.000 & 0.048 \\
A-OKVQA $K{=}8$ & 8B & 0.880 & 0.887 & +0.67 [-1.333, +3.000] & 0.080 & 0.000 & 0.047 \\
A-OKVQA $K{=}8$ & 7B & 0.867 & 0.873 & +0.67 [-1.000, +2.333] & 0.070 & 0.000 & 0.046 \\
A-OKVQA $K{=}16$ & 2B & 0.817 & 0.830 & +1.33 [-1.667, +4.333] & 0.130 & 0.000 & 0.091 \\
A-OKVQA $K{=}16$ & 4B & 0.817 & 0.840 & +2.33 [+0.333, +4.667] & 0.113 & 0.000 & 0.058 \\
A-OKVQA $K{=}16$ & 8B & 0.850 & 0.877 & +2.67 [+0.000, +5.333] & 0.103 & 0.000 & 0.075 \\
A-OKVQA $K{=}16$ & 7B & 0.830 & 0.853 & +2.33 [+0.000, +4.667] & 0.107 & 0.000 & 0.057 \\
EuroSAT & 2B & 0.521 & 0.524 & +0.30 [-1.800, +2.500] & 0.360 & 0.000 & 0.178 \\
EuroSAT & 4B & 0.573 & 0.555 & -1.80 [-3.800, +0.100] & 0.312 & 0.000 & 0.136 \\
EuroSAT & 8B & 0.540 & 0.547 & +0.70 [-1.000, +2.400] & 0.264 & 0.000 & 0.147 \\
EuroSAT & 7B & 0.379 & 0.382 & +0.30 [-1.400, +2.200] & 0.221 & 0.001 & 0.134 \\
Pets & 2B & 0.757 & 0.808 & +5.10 [+3.300, +7.000] & 0.259 & 0.000 & 0.130 \\
Pets & 4B & 0.890 & 0.893 & +0.30 [-1.000, +1.600] & 0.092 & 0.000 & 0.047 \\
Pets & 8B & 0.876 & 0.905 & +2.90 [+1.400, +4.400] & 0.124 & 0.000 & 0.058 \\
Pets & 7B & 0.921 & 0.931 & +1.00 [+0.198, +1.800] & 0.035 & 0.000 & 0.029 \\
MMMU & 2B & 0.479 & 0.497 & +1.78 [-1.065, +4.734] & 0.453 & 0.000 & 0.218 \\
MMMU & 4B & 0.549 & 0.581 & +3.20 [+0.473, +6.036] & 0.335 & 0.001 & 0.199 \\
MMMU & 8B & 0.560 & 0.583 & +2.37 [-0.473, +4.970] & 0.325 & 0.000 & 0.198 \\
MMMU & 7B & 0.522 & 0.541 & +1.89 [-0.710, +4.379] & 0.305 & 0.009 & 0.113 \\
\bottomrule
\end{tabular}
\caption{Canonical-order grid. L0 rotates the text-sorted listing; raw still uses the caller's listing. Correction TV is the mean TV between the single sorted read and its cycle average on the same listing.  }
\label{tab:app-canonical}
\end{table*}

\begin{table*}[t]
\centering
\footnotesize
\setlength{\tabcolsep}{3pt}
\begin{tabular}{@{}llrrrrr@{}}
\toprule
Dataset & Model & L0 & L0+batch & pointwise & pw+batch & grey $\alpha{=}1$ \\
\midrule
A-OKVQA & 2B & 0.834 & 0.834 & 0.829 & 0.829 & 0.743 \\
A-OKVQA & 4B & 0.866 & 0.866 & 0.845 & 0.845 & 0.790 \\
A-OKVQA & 8B & 0.879 & 0.879 & 0.867 & 0.867 & 0.825 \\
A-OKVQA & 7B & 0.881 & 0.881 & 0.864 & 0.864 & 0.795 \\
ScienceQA & 2B & 0.879 & 0.879 & 0.834 & 0.834 & 0.569 \\
ScienceQA & 4B & 0.930 & 0.929 & 0.909 & 0.909 & 0.712 \\
ScienceQA & 8B & 0.956 & 0.956 & 0.933 & 0.933 & 0.786 \\
ScienceQA & 7B & 0.882 & 0.883 & 0.874 & 0.873 & 0.726 \\
CV-Bench & 2B & 0.802 & 0.803 & 0.780 & 0.779 & 0.697 \\
CV-Bench & 4B & 0.854 & 0.854 & 0.835 & 0.835 & 0.838 \\
CV-Bench & 8B & 0.857 & 0.857 & 0.847 & 0.847 & 0.852 \\
CV-Bench & 7B & 0.792 & 0.792 & 0.763 & 0.762 & 0.697 \\
A-OKVQA $K{=}8$ & 2B & 0.833 & 0.833 & 0.807 & 0.807 & 0.617 \\
A-OKVQA $K{=}8$ & 4B & 0.860 & 0.860 & 0.840 & 0.840 & 0.740 \\
A-OKVQA $K{=}8$ & 8B & 0.880 & 0.880 & 0.853 & 0.853 & 0.780 \\
A-OKVQA $K{=}8$ & 7B & 0.870 & 0.870 & 0.827 & 0.827 & 0.690 \\
A-OKVQA $K{=}16$ & 2B & 0.817 & 0.817 & 0.770 & 0.770 & 0.577 \\
A-OKVQA $K{=}16$ & 4B & 0.847 & 0.847 & 0.820 & 0.820 & 0.700 \\
A-OKVQA $K{=}16$ & 8B & 0.867 & 0.867 & 0.830 & 0.830 & 0.763 \\
A-OKVQA $K{=}16$ & 7B & 0.860 & 0.860 & 0.803 & 0.803 & 0.630 \\
EuroSAT & 2B & 0.535 & 0.562 & 0.524 & 0.575 & 0.629 \\
EuroSAT & 4B & 0.545 & 0.557 & 0.557 & 0.564 & 0.522 \\
EuroSAT & 8B & 0.546 & 0.559 & 0.538 & 0.536 & 0.508 \\
EuroSAT & 7B & 0.366 & 0.411 & 0.355 & 0.389 & 0.331 \\
Pets & 2B & 0.808 & 0.830 & 0.818 & 0.849 & 0.835 \\
Pets & 4B & 0.893 & 0.901 & 0.900 & 0.910 & 0.910 \\
Pets & 8B & 0.905 & 0.908 & 0.899 & 0.910 & 0.901 \\
Pets & 7B & 0.931 & 0.939 & 0.925 & 0.935 & 0.924 \\
MMMU & 2B & 0.489 & 0.489 & 0.417 & 0.417 & 0.334 \\
MMMU & 4B & 0.574 & 0.574 & 0.498 & 0.498 & 0.391 \\
MMMU & 8B & 0.574 & 0.574 & 0.536 & 0.536 & 0.444 \\
MMMU & 7B & 0.531 & 0.531 & 0.485 & 0.485 & 0.400 \\
\bottomrule
\end{tabular}
\caption{Method comparison per cell. L0+batch is the per-question prior; where the prior is not estimable it equals L0. Grey adds the content-free prior to pointwise at the fixed AnyJev 0.3.0 default $\alpha{=}1$.  }
\label{tab:app-methods}
\end{table*}

\begin{table}[t]
\centering
\footnotesize
\setlength{\tabcolsep}{3pt}
\begin{tabular}{@{}llrrrr@{}}
\toprule
Dataset & Model & orig raw/L0 & rand raw/L0 & $\Delta$ orig & $\Delta$ rand \\
\midrule
EuroSAT & 7B & 0.353/0.343 & 0.347/0.357 & -1.0 & +1.0 \\
EuroSAT & 2B & 0.500/0.517 & 0.513/0.510 & +1.7 & -0.3 \\
EuroSAT & 4B & 0.547/0.537 & 0.510/0.530 & -1.0 & +2.0 \\
EuroSAT & 8B & 0.517/0.533 & 0.530/0.547 & +1.7 & +1.7 \\
Pets & 2B & 0.720/0.767 & 0.737/0.773 & +4.7 & +3.7 \\
Pets & 4B & 0.893/0.890 & 0.890/0.900 & -0.3 & +1.0 \\
\bottomrule
\end{tabular}
\caption{Slot randomization: the class-to-slot mapping is permuted per item. $\Delta$ is L0$-$raw accuracy in pp on 300-item subsets.  }
\label{tab:app-slot}
\end{table}

\begin{table}[t]
\centering
\scriptsize
\setlength{\tabcolsep}{4pt}
\begin{tabular}{@{}llrrr@{}}
\toprule
Dataset & Model & raw ECE & L0 ECE & change \\
\midrule
ScienceQA & 2B & 0.087 & 0.077 & -0.010 \\
ScienceQA & 4B & 0.062 & 0.058 & -0.005 \\
ScienceQA & 8B & 0.046 & 0.034 & -0.012 \\
ScienceQA & 7B & 0.023 & 0.019 & -0.003 \\
CV-Bench & 2B & 0.108 & 0.098 & -0.010 \\
CV-Bench & 4B & 0.119 & 0.112 & -0.007 \\
CV-Bench & 8B & 0.108 & 0.105 & -0.003 \\
CV-Bench & 7B & 0.030 & 0.032 & +0.002 \\
A-OKVQA $K{=}8$ & 2B & 0.141 & 0.129 & -0.012 \\
A-OKVQA $K{=}8$ & 4B & 0.123 & 0.117 & -0.006 \\
A-OKVQA $K{=}8$ & 8B & 0.108 & 0.101 & -0.007 \\
A-OKVQA $K{=}8$ & 7B & 0.050 & 0.043 & -0.006 \\
A-OKVQA $K{=}16$ & 2B & 0.141 & 0.129 & -0.013 \\
A-OKVQA $K{=}16$ & 4B & 0.167 & 0.143 & -0.024 \\
A-OKVQA $K{=}16$ & 8B & 0.141 & 0.115 & -0.026 \\
A-OKVQA $K{=}16$ & 7B & 0.076 & 0.052 & -0.023 \\
EuroSAT & 2B & 0.257 & 0.206 & -0.051 \\
EuroSAT & 4B & 0.343 & 0.368 & +0.025 \\
EuroSAT & 8B & 0.384 & 0.366 & -0.018 \\
EuroSAT & 7B & 0.186 & 0.199 & +0.013 \\
Pets & 2B & 0.187 & 0.130 & -0.057 \\
Pets & 4B & 0.099 & 0.095 & -0.004 \\
Pets & 8B & 0.110 & 0.077 & -0.032 \\
Pets & 7B & 0.021 & 0.016 & -0.005 \\
MMMU & 2B & 0.303 & 0.239 & -0.064 \\
MMMU & 4B & 0.332 & 0.282 & -0.050 \\
MMMU & 8B & 0.333 & 0.286 & -0.048 \\
MMMU & 7B & 0.122 & 0.095 & -0.028 \\
\bottomrule
\end{tabular}
\caption{Expected calibration error (10 equal-width bins) before and after L0. Negative change means L0 is better calibrated.  }
\label{tab:app-ece}
\end{table}

\begin{table*}[t]
\centering
\footnotesize
\setlength{\tabcolsep}{2.5pt}
\begin{tabular}{@{}llrrrr@{}}
\toprule
Dataset & Model & flip raw [95\% CI] & flip L0 [95\% CI] & TV raw [95\% CI] & TV L0 [95\% CI] \\
\midrule
ScienceQA & 2B & 0.112 [0.098, 0.126] & 0.024 [0.018, 0.031] & 0.100 [0.090, 0.110] & 0.026 [0.022, 0.030] \\
ScienceQA & 4B & 0.059 [0.049, 0.069] & 0.018 [0.012, 0.024] & 0.057 [0.049, 0.066] & 0.018 [0.014, 0.023] \\
ScienceQA & 8B & 0.049 [0.040, 0.059] & 0.013 [0.009, 0.019] & 0.050 [0.041, 0.058] & 0.011 [0.008, 0.015] \\
ScienceQA & 7B & 0.090 [0.077, 0.102] & 0.020 [0.014, 0.027] & 0.072 [0.065, 0.078] & 0.016 [0.014, 0.018] \\
CV-Bench & 2B & 0.127 [0.114, 0.139] & 0.019 [0.014, 0.025] & 0.109 [0.102, 0.116] & 0.013 [0.011, 0.015] \\
CV-Bench & 4B & 0.063 [0.055, 0.073] & 0.014 [0.009, 0.018] & 0.059 [0.053, 0.066] & 0.012 [0.010, 0.014] \\
CV-Bench & 8B & 0.053 [0.044, 0.061] & 0.012 [0.008, 0.017] & 0.046 [0.041, 0.052] & 0.010 [0.009, 0.012] \\
CV-Bench & 7B & 0.073 [0.063, 0.083] & 0.018 [0.013, 0.023] & 0.068 [0.065, 0.071] & 0.007 [0.007, 0.008] \\
A-OKVQA $K{=}8$ & 2B & 0.090 [0.060, 0.123] & 0.037 [0.017, 0.060] & 0.093 [0.069, 0.117] & 0.036 [0.026, 0.046] \\
A-OKVQA $K{=}8$ & 4B & 0.073 [0.047, 0.103] & 0.040 [0.020, 0.063] & 0.070 [0.047, 0.095] & 0.035 [0.021, 0.051] \\
A-OKVQA $K{=}8$ & 8B & 0.080 [0.050, 0.113] & 0.040 [0.020, 0.063] & 0.079 [0.054, 0.105] & 0.035 [0.021, 0.050] \\
A-OKVQA $K{=}8$ & 7B & 0.070 [0.043, 0.100] & 0.037 [0.017, 0.057] & 0.080 [0.063, 0.097] & 0.033 [0.026, 0.041] \\
A-OKVQA $K{=}16$ & 2B & 0.130 [0.093, 0.167] & 0.040 [0.020, 0.063] & 0.133 [0.103, 0.163] & 0.048 [0.035, 0.062] \\
A-OKVQA $K{=}16$ & 4B & 0.113 [0.080, 0.150] & 0.033 [0.013, 0.057] & 0.099 [0.072, 0.129] & 0.039 [0.025, 0.055] \\
A-OKVQA $K{=}16$ & 8B & 0.103 [0.070, 0.140] & 0.027 [0.010, 0.047] & 0.101 [0.072, 0.132] & 0.037 [0.023, 0.052] \\
A-OKVQA $K{=}16$ & 7B & 0.107 [0.073, 0.140] & 0.043 [0.020, 0.067] & 0.099 [0.080, 0.119] & 0.038 [0.030, 0.047] \\
EuroSAT & 2B & 0.360 [0.332, 0.390] & 0.291 [0.263, 0.319] & 0.331 [0.311, 0.352] & 0.225 [0.213, 0.237] \\
EuroSAT & 4B & 0.312 [0.284, 0.339] & 0.068 [0.054, 0.084] & 0.300 [0.277, 0.321] & 0.076 [0.067, 0.085] \\
EuroSAT & 8B & 0.264 [0.237, 0.292] & 0.127 [0.106, 0.148] & 0.249 [0.228, 0.271] & 0.108 [0.096, 0.119] \\
EuroSAT & 7B & 0.221 [0.196, 0.248] & 0.121 [0.101, 0.142] & 0.195 [0.187, 0.203] & 0.111 [0.107, 0.116] \\
Pets & 2B & 0.259 [0.232, 0.287] & 0.106 [0.088, 0.126] & 0.252 [0.227, 0.276] & 0.101 [0.088, 0.115] \\
Pets & 4B & 0.092 [0.074, 0.111] & 0.064 [0.049, 0.079] & 0.092 [0.075, 0.108] & 0.065 [0.052, 0.078] \\
Pets & 8B & 0.124 [0.104, 0.144] & 0.076 [0.060, 0.093] & 0.124 [0.105, 0.143] & 0.076 [0.063, 0.091] \\
Pets & 7B & 0.035 [0.024, 0.046] & 0.033 [0.023, 0.044] & 0.053 [0.047, 0.060] & 0.036 [0.032, 0.040] \\
MMMU & 2B & 0.453 [0.420, 0.488] & 0.201 [0.175, 0.228] & 0.349 [0.329, 0.370] & 0.124 [0.114, 0.134] \\
MMMU & 4B & 0.335 [0.304, 0.366] & 0.156 [0.133, 0.181] & 0.310 [0.286, 0.334] & 0.113 [0.102, 0.124] \\
MMMU & 8B & 0.325 [0.295, 0.356] & 0.140 [0.117, 0.163] & 0.293 [0.270, 0.317] & 0.113 [0.102, 0.125] \\
MMMU & 7B & 0.305 [0.276, 0.337] & 0.150 [0.127, 0.175] & 0.183 [0.172, 0.194] & 0.057 [0.053, 0.061] \\
\bottomrule
\end{tabular}
\caption{Per-cell flip and aligned-TV intervals (item bootstrap, 2,000--4,000 resamples). These are the uncertainty estimates behind the point estimates in Table~\ref{tab:app-stage2}; no multiplicity correction is applied.  }
\label{tab:app-interval}
\end{table*}

\begin{table}[t]
\centering
\footnotesize
\setlength{\tabcolsep}{4pt}
\begin{tabular}{@{}llrrr@{}}
\toprule
Dataset & Model & raw & L0 & change \\
\midrule
ScienceQA & 2B & 0.794 & 0.814 & +0.020 \\
ScienceQA & 4B & 0.937 & 0.966 & +0.029 \\
ScienceQA & 8B & 0.995 & 1.000 & +0.005 \\
ScienceQA & 7B & 0.831 & 0.853 & +0.022 \\
CV-Bench & 2B & 0.510 & 0.517 & +0.007 \\
CV-Bench & 4B & 0.701 & 0.684 & -0.016 \\
CV-Bench & 8B & 0.761 & 0.779 & +0.018 \\
CV-Bench & 7B & 0.293 & 0.281 & -0.012 \\
A-OKVQA $K{=}8$ & 2B & 0.650 & 0.617 & -0.033 \\
A-OKVQA $K{=}8$ & 4B & 0.757 & 0.753 & -0.003 \\
A-OKVQA $K{=}8$ & 8B & 0.780 & 0.753 & -0.027 \\
A-OKVQA $K{=}8$ & 7B & 0.793 & 0.833 & +0.040 \\
A-OKVQA $K{=}16$ & 2B & 0.567 & 0.553 & -0.013 \\
A-OKVQA $K{=}16$ & 4B & 0.700 & 0.733 & +0.033 \\
A-OKVQA $K{=}16$ & 8B & 0.750 & 0.773 & +0.023 \\
A-OKVQA $K{=}16$ & 7B & 0.780 & 0.803 & +0.023 \\
EuroSAT & 2B & 0.216 & 0.207 & -0.009 \\
EuroSAT & 4B & 0.182 & 0.222 & +0.040 \\
EuroSAT & 8B & 0.015 & 0.030 & +0.015 \\
EuroSAT & 7B & 0.002 & 0.002 & +0.000 \\
Pets & 2B & 0.474 & 0.656 & +0.182 \\
Pets & 4B & 0.848 & 0.867 & +0.019 \\
Pets & 8B & 0.822 & 0.858 & +0.036 \\
Pets & 7B & 0.936 & 0.951 & +0.015 \\
MMMU & 2B & 0.031 & 0.014 & -0.017 \\
MMMU & 4B & 0.178 & 0.191 & +0.013 \\
MMMU & 8B & 0.124 & 0.121 & -0.004 \\
MMMU & 7B & 0.066 & 0.057 & -0.009 \\
\bottomrule
\end{tabular}
\caption{Coverage at 5\% selective risk before and after L0. The threshold is selected on the same items, so this is an optimistic operating point.  }
\label{tab:app-coverage}
\end{table}

\begin{table*}[t]
\centering
\scriptsize
\setlength{\tabcolsep}{4pt}
\begin{tabular}{@{}llrrrr@{}}
\toprule
Family & Set & $n$ & raw & L0 & flip raw $\to$ L0 \\
\midrule
LLaVA-OneVision-7B & A-OKVQA & 1145 & 0.9135 & 0.9205 & 0.038 $\to$ 0.015 \\
LLaVA-OneVision-7B & ScienceQA & 2017 & 0.9593 & 0.9618 & 0.027 $\to$ 0.006 \\
LLaVA-OneVision-7B & EuroSAT & 1000 & 0.4870 & 0.4660 & 0.571 $\to$ 0.076 \\
LLaVA-OneVision-7B & Pets & 1000 & 0.6940 & 0.7100 & 0.138 $\to$ 0.085 \\
LLaVA-NeXT-7B & A-OKVQA & 1145 & 0.8166 & 0.8384 & 0.105 $\to$ 0.057 \\
LLaVA-NeXT-7B & ScienceQA & 2017 & 0.7234 & 0.7224 & 0.208 $\to$ 0.076 \\
LLaVA-NeXT-7B & EuroSAT & 1000 & 0.4930 & 0.5050 & 0.598 $\to$ 0.309 \\
LLaVA-NeXT-7B & Pets & 1000 & 0.4440 & 0.4900 & 0.554 $\to$ 0.414 \\
Idefics3-8B & A-OKVQA & 1145 & 0.8201 & 0.8306 & 0.135 $\to$ 0.045 \\
Idefics3-8B & ScienceQA & 2017 & 0.8825 & 0.8909 & 0.060 $\to$ 0.009 \\
Idefics3-8B & EuroSAT & 1000 & 0.3560 & 0.3970 & 0.292 $\to$ 0.136 \\
Idefics3-8B & Pets & 1000 & 0.6110 & 0.6860 & 0.411 $\to$ 0.088 \\
\bottomrule
\end{tabular}
\caption{Full non-Qwen cells. $n$ varies by dataset and is shown per row; EuroSAT cells use the 1,000-item subset (the Idefics3 full-split 2,700 run is noted in the text). Accuracies are rounded to four decimals.}
\label{tab:app-full}
\end{table*}

\begin{table}[t]
\centering
\scriptsize
\setlength{\tabcolsep}{4pt}
\begin{tabular}{@{}lrrrr@{}}
\toprule
Model & raw & L0 & flip raw & flip L0 \\
\midrule
Qwen2.5-VL-7B-Instruct & 0.793 & 0.817 & 0.170 & 0.087 \\
Qwen3-VL-2B-Instruct & 0.757 & 0.760 & 0.217 & 0.110 \\
Qwen3-VL-4B-Instruct & 0.727 & 0.780 & 0.183 & 0.087 \\
Qwen3-VL-8B-Instruct & 0.817 & 0.810 & 0.140 & 0.073 \\
\bottomrule
\end{tabular}
\caption{BANKING77 text-only control with the same VLM readout (20 intents, 300 items).  }
\label{tab:app-text}
\end{table}

\begin{table}[t]
\centering
\scriptsize
\setlength{\tabcolsep}{3pt}
\begin{tabular}{@{}llrrr@{}}
\toprule
Set / model & Variant & raw & L0 & TV \\
\midrule
A-OKVQA 7B & original & 0.890 & 0.900 & 0.027 \\
A-OKVQA 7B & case & 0.890 & 0.900 & 0.027 \\
A-OKVQA 7B & reverse\_words & 0.865 & 0.870 & 0.031 \\
A-OKVQA 2B & original & 0.850 & 0.870 & 0.038 \\
A-OKVQA 2B & case & 0.850 & 0.870 & 0.038 \\
A-OKVQA 2B & reverse\_words & 0.850 & 0.830 & 0.042 \\
A-OKVQA 4B & original & 0.875 & 0.880 & 0.035 \\
A-OKVQA 4B & case & 0.875 & 0.880 & 0.035 \\
A-OKVQA 4B & reverse\_words & 0.840 & 0.840 & 0.038 \\
A-OKVQA 8B & original & 0.880 & 0.905 & 0.032 \\
A-OKVQA 8B & case & 0.880 & 0.905 & 0.032 \\
A-OKVQA 8B & reverse\_words & 0.845 & 0.880 & 0.038 \\
EuroSAT 2B & original & 0.520 & 0.545 & 0.182 \\
EuroSAT 2B & case & 0.520 & 0.545 & 0.182 \\
EuroSAT 2B & synonym & 0.500 & 0.490 & 0.107 \\
EuroSAT 2B & reverse\_words & 0.565 & 0.520 & 0.241 \\
EuroSAT 4B & original & 0.530 & 0.525 & 0.123 \\
EuroSAT 4B & case & 0.530 & 0.525 & 0.123 \\
EuroSAT 4B & synonym & 0.470 & 0.455 & 0.130 \\
EuroSAT 4B & reverse\_words & 0.540 & 0.485 & 0.103 \\
\bottomrule
\end{tabular}
\caption{Option-text paraphrases. \texttt{case} lowercases; \texttt{reverse\_words} reverses word order; \texttt{synonym} uses the EuroSAT mapping.  }
\label{tab:app-paraphrase}
\end{table}

\begin{table}[t]
\centering
\scriptsize
\setlength{\tabcolsep}{4pt}
\begin{tabular}{@{}llrrr@{}}
\toprule
Set & Model & no prior & held-out & in-sample \\
\midrule
EuroSAT & 2B & 0.534 & 0.566 & 0.564 \\
EuroSAT & 4B & 0.514 & 0.526 & 0.528 \\
EuroSAT & 8B & 0.544 & 0.556 & 0.558 \\
EuroSAT & 7B & 0.340 & 0.380 & 0.388 \\
Pets & 2B & 0.826 & 0.844 & 0.848 \\
Pets & 4B & 0.896 & 0.902 & 0.900 \\
Pets & 8B & 0.922 & 0.924 & 0.924 \\
Pets & 7B & 0.938 & 0.944 & 0.948 \\
\bottomrule
\end{tabular}
\caption{Fixed-question prior with a held-out prefix: no prior, prior estimated on half the items and applied to the other half, and the transductive in-sample result on the same test half.  }
\label{tab:app-heldout}
\end{table}

\begin{table}[t]
\centering
\scriptsize
\setlength{\tabcolsep}{4pt}
\begin{tabular}{@{}lrrrr@{}}
\toprule
Model & orig raw/L0 & rand raw/L0 & $\Delta$ orig & $\Delta$ rand \\
\midrule
EuroSAT 2B & 0.521/0.535 & 0.542/0.549 & +1.4 & +0.7 \\
EuroSAT 4B & 0.573/0.545 & 0.551/0.545 & -2.8 & -0.6 \\
Pets 2B & 0.757/0.808 & 0.775/0.810 & +5.1 & +3.5 \\
Pets 4B & 0.890/0.893 & 0.888/0.891 & +0.3 & +0.3 \\
EuroSAT/ID3 8B & 0.356/0.397 & 0.330/0.374 & +4.1 & +4.4 \\
\bottomrule
\end{tabular}
\caption{Slot randomization on 1,000-item sets (EuroSAT/Qwen, Pets/Qwen, EuroSAT/Idefics3). $\Delta$ is L0$-$raw accuracy in pp.  }
\label{tab:app-slot-full}
\end{table}

\begin{table}[t]
\centering
\scriptsize
\setlength{\tabcolsep}{4pt}
\begin{tabular}{@{}llrrr@{}}
\toprule
Set & Model & base & context & context+batch \\
\midrule
A-OKVQA full & 2B & 0.829 & 0.808 & 0.808 \\
A-OKVQA full & 4B & 0.845 & 0.852 & 0.852 \\
EuroSAT & 2B & 0.525 & 0.530 & 0.570 \\
EuroSAT & 4B & 0.555 & 0.505 & 0.525 \\
\bottomrule
\end{tabular}
\caption{Options-context pointwise control: A-OKVQA full 1145, EuroSAT 200. The full option list is visible, but one option is scored.  }
\label{tab:app-context}
\end{table}

\begin{table*}[t]
\centering
\scriptsize
\setlength{\tabcolsep}{4pt}
\begin{tabular}{@{}llrrrr@{}}
\toprule
Family & Set & $n$ & L0 & pointwise & $\Delta$ \\
\midrule
LLaVA-OneVision-7B & A-OKVQA & 1145 & 0.921 & 0.889 & -3.1 \\
LLaVA-OneVision-7B & ScienceQA & 2017 & 0.962 & 0.952 & -1.0 \\
LLaVA-OneVision-7B & EuroSAT & 1000 & 0.466 & 0.521 & +5.5 \\
LLaVA-OneVision-7B & Pets & 1000 & 0.710 & 0.673 & -3.7 \\
LLaVA-NeXT-7B & A-OKVQA & 1145 & 0.838 & 0.803 & -3.5 \\
LLaVA-NeXT-7B & ScienceQA & 2017 & 0.722 & 0.720 & -0.2 \\
LLaVA-NeXT-7B & EuroSAT & 1000 & 0.505 & 0.537 & +3.2 \\
LLaVA-NeXT-7B & Pets & 1000 & 0.490 & 0.493 & +0.3 \\
Idefics3-8B & A-OKVQA & 1145 & 0.831 & 0.828 & -0.3 \\
Idefics3-8B & ScienceQA & 2017 & 0.891 & 0.875 & -1.6 \\
Idefics3-8B & EuroSAT & 1000 & 0.397 & 0.323 & -7.4 \\
Idefics3-8B & Pets & 1000 & 0.686 & 0.743 & +5.7 \\
\bottomrule
\end{tabular}
\caption{Full-cell pointwise accuracy against the matched full L0 cell. $\Delta$ is pointwise$-$L0 in pp. EuroSAT uses the 1,000-item subset.  }
\label{tab:app-pointwise-full}
\end{table*}

\begin{table}[t]
\centering
\scriptsize
\setlength{\tabcolsep}{3pt}
\begin{tabular}{@{}lrrrrr@{}}
\toprule
Set & raw & canonical L0 & $\Delta$ & flip raw $\to$ L0 & corr.\ TV \\
\midrule
A-OKVQA & 0.817 & 0.834 & +1.7 & 0.105 $\to$ 0.001 & 0.060 \\
ScienceQA & 0.723 & 0.714 & -0.9 & 0.208 $\to$ 0.008 & 0.067 \\
EuroSAT & 0.493 & 0.530 & +3.7 & 0.598 $\to$ 0.000 & 0.286 \\
Pets & 0.444 & 0.490 & +4.6 & 0.554 $\to$ 0.000 & 0.244 \\
\bottomrule
\end{tabular}
\caption{LLaVA-NeXT full protocol: canonical-order L0. $\Delta$ is canonical L0$-$raw in pp; correction TV is the distance between the single sorted read and its cycle average.  }
\label{tab:app-next-canonical}
\end{table}

\begin{table}[t]
\centering
\scriptsize
\setlength{\tabcolsep}{3pt}
\begin{tabular}{@{}lrrrrr@{}}
\toprule
Set & L0 & L0+prior & pointwise & +grey $\alpha{=}1$ & +options \\
\midrule
A-OKVQA & 0.838 & -- & 0.803 & -- & 0.775 \\
ScienceQA & 0.722 & -- & 0.720 & -- & 0.706 \\
EuroSAT & 0.505 & 0.581 & 0.537 & 0.538 & 0.378 \\
Pets & 0.490 & 0.547 & 0.493 & 0.481 & 0.368 \\
\bottomrule
\end{tabular}
\caption{LLaVA-NeXT full protocol: priors, grey, and options-context controls. The prior is estimable only on Pets/EuroSAT; grey is pointwise plus the content-free prior at $\alpha{=}1$; +options is pointwise with the full option list visible. }
\label{tab:app-next-protocol}
\end{table}

\begin{table*}[t]
\centering
\scriptsize
\setlength{\tabcolsep}{4pt}
\begin{tabular}{@{}llrrrr@{}}
\toprule
Model & Set & grey $H$ & top-1 match & overlap & $\Delta\alpha{=}1$ \\
\midrule
Qwen 7B & EuroSAT & 3.00 & no & 0.758 & -2.4 \\
Qwen 2B & EuroSAT & 2.32 & yes & 0.758 & +10.5 \\
Qwen 4B & EuroSAT & 0.73 & yes & 0.303 & -3.5 \\
Qwen 8B & EuroSAT & 0.93 & no & 0.342 & -3.0 \\
Qwen 7B & Pets & 4.58 & no & 0.820 & -0.1 \\
Qwen 2B & Pets & 4.02 & yes & 0.626 & +1.7 \\
Qwen 4B & Pets & 3.75 & no & 0.567 & +1.0 \\
Qwen 8B & Pets & 4.18 & no & 0.638 & +0.2 \\
LLaVA-NeXT & EuroSAT & 3.28 & no & 0.956 & +0.1 \\
LLaVA-NeXT & Pets & 4.57 & no & 0.832 & -1.2 \\
\bottomrule
\end{tabular}
\caption{Grey-prior diagnostic on fixed-option cells. $H$ is the entropy of the mean grey marginal; top-1 match compares its argmax to the real pointwise marginal; overlap is $\sum_c\min(g_c,r_c)$; $\Delta$ is pointwise accuracy change at $\alpha{=}1$ in pp.  }
\label{tab:app-grey-diagnostic}
\end{table*}

\subsection{Additional ablations}\label{sec:app_ablations}

\textbf{Arbitrary permutations.} Reversal is one alternative ordering; arbitrary permutations are a stronger
probe~\citep{zong2024fool,paruchuri2026same}.  On 300-item subsets, five random permutations
per item change 5.0\% of A-OKVQA answers, 24.4\% on EuroSAT and 10.6\% on Pets,
relative to the raw read. All five permutations return the raw answer on
90\%, 53\% and 78\% of items respectively. Log-averaging the five permutations
matches cyclic L0 within 0.8 points of accuracy, so the cyclic family is a good
proxy for arbitrary ordering.

\textbf{Prompt formats.} Prompt formats move accuracy much less.  Replacing the default \texttt{A.} delimiter with
\texttt{A)}, \texttt{A:} or \texttt{Option A:} moves mean accuracy by at most
2.3 points on A-OKVQA (2B) and at most 1.7 points on the other models; raw
reversal flips stay at 0.047--0.087 and L0 flips at 0.010--0.037. The
conclusions are not a delimiter artifact.  

\textbf{Pointwise batch size.} On 300 items with Qwen3-VL-2B, pointwise
accuracy is 0.827 at batch 1 and 0.830 at batch 4 on A-OKVQA, and 0.817 vs 0.807
on Pets. Batching changes the forward composition, not the readout definition.

\textbf{Batch-prior activation.} The per-question prior activates on 993 of
1,000 items on the fixed-question sets and on 0 items for A-OKVQA, the synthetic
$K{=}8/16$ sets and MMMU; ScienceQA activates 65 of 2,017 and CV-Bench 44 of
2,638, where the effect is at most 0.5 points. These counts are in the prior
runner's per-cell JSON.

\paragraph{Text-domain control.}
The same VLM readout and L0 were run on BANKING77 (300 items, 20 intents,
text-only state). Accuracy raw/L0 is 0.757/0.760 (2B), 0.727/0.780 (4B),
0.817/0.810 (8B) and 0.793/0.817 (7B); reversal flip falls from 0.217 to 0.110,
0.183 to 0.087, 0.140 to 0.073 and 0.170 to 0.087. Text-domain L0 therefore
behaves like the visual readout: it roughly halves the flip and changes accuracy
modestly. The visual findings are not an artifact of the letter readout failing
on images. Table~\ref{tab:app-text}.

\paragraph{Option paraphrases.}
Lowercasing the option text leaves every result unchanged (the tokenizer is
case-insensitive here). Reversing word order is not benign: A-OKVQA L0 drops by
2.5--4 points across the four models. EuroSAT synonyms are larger: L0 falls
0.545$\to$0.490 (2B) and 0.525$\to$0.455 (4B), while reverse word order gives
0.520 and 0.485. Raw accuracy can move in the other direction, so this is not
just an L0 failure. Option wording is a larger perturbation than the delimiter
styles, and L0 does not remove it. Table~\ref{tab:app-paraphrase}.

\subsection{Statistical and scope caveats}
\label{sec:app-caveats}

The Qwen grid has four models; full-cell cross-family evaluation agrees on
flips, not pointwise, and only LLaVA-NeXT receives the full non-Qwen method
protocol. Datasets are English and short; large-$K$ evidence is Pets and
synthetic $K{=}16$. Grey ``best $\alpha$ is selected on test and optimistic;
$\alpha{=}1$ is the honest fixed choice. Pointwise invariance is exact only at
batch 1; the fixed-question prior is transductive. We sample five permutations
and four delimiters; ScienceQA’s hint survives grey. One GPU, bfloat16; we
measure decisions, not downstream utility.

The main grid is 28 cells per readout; we do not apply a family-wise
multiple-comparison correction. The item-level intervals are descriptive, and
the class-cluster intervals for Pets/EuroSAT are the conservative read. The
reduced non-Qwen grids use 300 items (Pets 200) and are underpowered by design;
the full non-Qwen cells in Table~\ref{tab:app-full} are reported separately.
On fixed-question sets the per-question prior is estimated over the test stream
and includes the current item, so part of its gain may be correcting class
imbalance rather than label bias; a deployment should use a held-out prefix.
A held-out-prefix approximation on Pets/EuroSAT retains gains of +0.2 to +4.0
points, versus +0.2 to +4.8 in-sample (Table~\ref{tab:app-heldout}); the
transductive component is small but nonzero.
The pointwise comparison also changes the prompt structure, not only the
presence of a contrast set; the contrast-set account is consistent with, not
proven by, these runs. A control that shows the full option list while scoring one option is mixed:
on full A-OKVQA, 2B changes by $-2.1$ points and 4B by $+0.7$; on the 200-item
EuroSAT subset, 2B changes by $+0.5$ and 4B by $-5.0$
(Table~\ref{tab:app-context}). Showing the other options does not consistently
close the pointwise gap.
On the 1,000-item EuroSAT evaluation set, slot randomization moves L0$-$raw from
$+1.4$ to $+0.7$ points at 2B and from $-2.8$ to $-0.6$ at 4B; the 4B
randomized interval includes zero (Table~\ref{tab:app-slot-full}).
The grey figure's ``best $\alpha$'' is selected on the same cell and is optimistic; the main table
uses the fixed package default, and the held-out half-split is the honest
estimate. The synthetic $K{=}8/16$ sets are stress tests, not natural large-$K$
benchmarks. Pointwise reads only single-token yes/no variants; the recorded
yes+no mass is 0.88--1.0 except CV-Bench 2B (0.42), which bounds how much
probability the readout can miss.
ScienceQA's hint text survives the grey probe, so its content-free readout is
image-content-free rather than fully content-free.

\begin{figure*}[t]
\centering
\includegraphics[width=\textwidth]{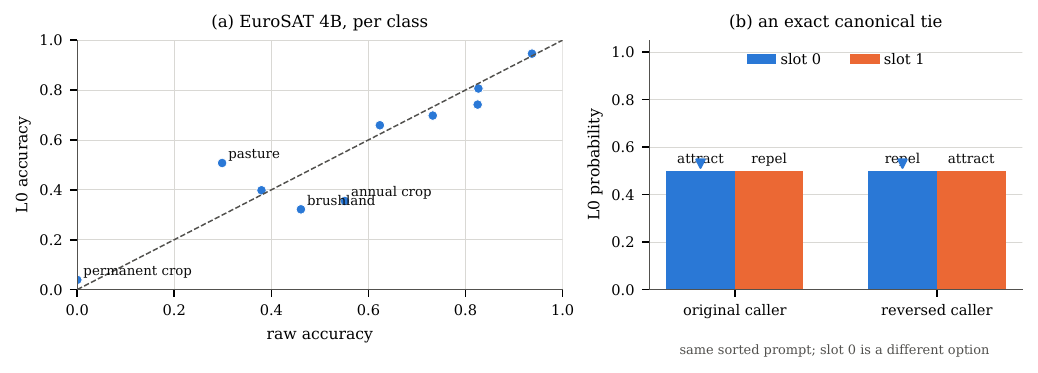}
\caption{(a) EuroSAT 4B per-class accuracy before and after L0; points below the
diagonal lose under L0. The fixed class list makes slot and class the same
variable, so rotation moves a class advantage. (b) The exact canonical tie from
\texttt{scienceqa\_80}: the sorted prompt is identical under reversal and the
aligned probabilities are both 0.5, but slot 0 names ``attract'' in the original
caller array and ``repel'' in the reversed one. The reported flip is the index
tie-break, not a different sorted listing.}
\label{fig:app-qual}
\end{figure*}

\paragraph{Selective prediction and an action proxy.}
At a 5\% selective-risk threshold, L0 raises mean coverage from 0.553 to 0.568
(17 of 28 cells improve; Pets +6.3, ScienceQA +1.9;
Table~\ref{tab:app-coverage}). We also treat the top-1
label as an action when its confidence passes that threshold. Among accepted
actions, reversal changes 1.6\% of raw actions and 0.3\% of L0 actions. A
reversal check (act only when the original and reversed reads agree) covers
84.1\% of items under raw and 93.0\% under L0; accuracy among agreeing items is
81.5\% and 79.3\%, so agreement is a stability filter, not a correctness
guarantee. The 5\% threshold is chosen on the same items, so this coverage is an
optimistic operating point. This is a decision-level proxy, not a user study.  

\paragraph{Cell-level mixed model.}
Treating the 28 cells as exchangeable within datasets, a linear mixed model
with a dataset random intercept and model fixed effects estimates accuracy
$+0.72$ points $[-0.39,+1.83]$ ($p{=}0.20$), flip reduction $+6.83$
$[+1.92,+11.73]$ ($p{=}0.006$) and TV reduction $+6.43$ $[+1.93,+10.93]$
($p{=}0.005$). There are only seven dataset clusters, so this is a descriptive
hierarchical summary, not a substitute for the per-cell intervals. Fit with
statsmodels 0.15.0.

\paragraph{Modality-token order is a separate study.}
Ordering the image and text parts is a different intervention from ordering the
options: it changes the prompt construction, not the label readout, and L0 over
option labels cannot correct it. Prior work already documents that modality
order matters~\citep{gan2026order,wei2026order}. A separate study should permute
the visual/text part order at a fixed option list and measure the same flip/TV
statistics; we deliberately do not fold it into the option-list claims here.

\subsection{Worked examples}

Figure~\ref{fig:app-qual} shows both examples.

\textbf{An exact canonical tie.} ScienceQA item \texttt{scienceqa\_80} has two options,
``attract'' and ``repel,'' and sorted L0 probabilities $[0.5,0.5]$. The sorted
prompt is identical under reversal, so the aligned distributions are
bit-identical. The original run reports slot 0 as ``attract''; the reversed
caller run reports slot 0 as ``repel.'' The flip is the argmax tie-break in the
caller's array, not a different sorted listing.

\textbf{The fixed-list confound.} EuroSAT always shows the same ten classes in
the same order: annual crop at slot 0, pasture at slot 5, and so on. The raw
position marginal over-predicts early slots (annual crop 0.176, forest 0.227;
true shares about 0.10) and suppresses pasture (0.031 versus 0.067). L0 removes
the position effect, which here also removes the class advantage; pasture rises
from 0.30 to 0.51 while annual crop falls from 0.55 to 0.36. Pets does not show
the same loss because its breed list is 26 visually distinctive classes and its
raw accuracy is high.  

\textbf{The MMMU option string.} An MMMU row may store
\texttt{options} as the string \texttt{"['\$6', '\$7']"}. Iterating that string
as if it were a list produces one-character options and near-chance accuracy.
The adapter calls \texttt{ast.literal\_eval} first; the corrected 845-item set
is the one used in this paper, and the old runs are quarantined.

\end{document}